%% file: _paper.tex
\documentclass[12pt,twoside]{article}
\usepackage{arxiv}

\include{tikz_header}

\usepackage{amssymb}
\usepackage{amsmath}
\usepackage{subfig}
\usepackage{bbding}
\usepackage[normalem]{ulem} 
\usepackage{booktabs}
\usepackage[draft]{minted}

\usepackage{pifont}
\newcommand{\xmark}{\ding{55}}
\definecolor{LightGray}{gray}{0.9}
\definecolor{darkblue}{HTML}{1f4e79}
\definecolor{lightblue}{HTML}{00b0f0}
\definecolor{salmon}{HTML}{ff9c6b}
\definecolor{lightgreen}{HTML}{90ee90}

\usepackage{lineno}
\usepackage[utf8]{inputenc}
\usepackage{listings}
\usepackage{url}
\usepackage[hidelinks]{hyperref}

\input{minted_preamble}

\usepackage{cleveref}

\title{TSFE$_{DL}$: A Python Library for Time Series Spatio-Temporal Feature Extraction and Prediction using Deep Learning (with Appendices on Detailed Network Architectures and Experimental Cases of Study)}

\author{Ignacio Aguilera-Martos (nacheteam@ugr.es)$^{1,2}$ \And Ángel M. García-Vico (agvico@decsai.ugr.es)$^{1,2}$ \And Julián Luengo (julianlm@decsai.ugr.es)$^{1,2}$ \And Sergio Damas (sdamas@ugr.es)$^{3,2}$ \And Francisco J. Melero (fjmelero@ugr.es)$^{3,2}$ \And José Javier Valle-Alonso $^{4}$ \And Francisco Herrera (herrera@decsai.ugr.es)$^{1,2}$ \\
Andalusian Institute of Data Science and Computational Intelligence (DaSCI)\\
Department of Computer Science and Artificial Intelligence, University of Granada, Granada, Spain\\
Department of Software Engineering, University of Granada, Granada, Spain\\
Repsol Technology Lab, Spain}

\date{June, 2022}

\hypersetup{
pdftitle={TSFEDL: A Python Library for Time Series Spatio-Temporal Feature Extraction and Prediction using Deep Learning (with Appendices on Detailed Network Architectures and Experimental Cases of Study)},
pdfsubject={Neural and Evolutionary Computing},
pdfauthor={Ignacio Aguilera-Martos, \'Angel M. Garc\'ia-Vico, Juli\'an Luengo, Sergio Damas, Francisco J. Melero, Jos\'e Javier Valle-Alonso, Francisco Herrera},
pdfkeywords={Time series, Deep Learning, Python},
hypertexnames=false,
}

\renewcommand{\headeright}{}
\renewcommand{\undertitle}{}
\begin{document}

\maketitle

\begin{abstract}

The combination of convolutional and recurrent neural networks is a promising framework that allows the extraction of high-quality spatio-temporal features together with its temporal dependencies, which is key for time series prediction problems such as forecasting, classification or anomaly detection, amongst others. In this paper, the TSFE$_{DL}$ library is introduced. It compiles 20 state-of-the-art methods for both time series feature extraction and prediction, employing convolutional and recurrent deep neural networks for its use in several data mining tasks. The library is built upon a set of \texttt{Tensorflow+Keras} and \texttt{PyTorch} modules under the \texttt{AGPLv3} license. The performance validation of the architectures included in this proposal confirms the usefulness of this Python package.

\end{abstract}


\section{Introduction}
\label{sec:introduction}

The success of Machine Learning algorithms for time series prediction problems depends on the quality of the spatio-temporal features extracted. Deep Learning \cite{deep-learning} models can produce non-linear transformations on data, yielding more abstract and useful spatio-temporal features and patterns than classical models for a better prediction. Among the different areas of Deep Learning, the combination of Convolutional Neural Networks (CNNs) and Recurrent Neural Networks (RNNs) is an important novelty for spatio-temporal feature extraction in time series \cite{Tan2018}. This symbiosis allows the extraction of high-quality spatial features using CNNs, whereas the RNNs find the temporal dependencies among them. 

This neural network paradigm is providing interesting results in several areas, focusing the initial application on arrhythmia detection \cite{OhShuLih,GaoJunli,ChenChen}, with applications on other medical areas \cite{WeiXiaoyan}, energy forecasting \cite{KhanZulfiqar,KimTaeYoung} or remaining useful life prediction \cite{KongZhengmin}. The absence of a unified collection of neural networks for time series generates the necessity of an easy to use and performant solution for the practitioner. The Python package \texttt{TSFE$_{DL}$}, presented in this paper, supports this process by providing a wide variety of easily customisable CNN-RNN Deep Learning models. All the available architectures are implemented from scratch, unifying the programming style and providing open source code for the included networks, presenting the user accessible and useful Python code.

The rest of the paper is structured as follows: \Cref{sec:software_framework} explains the software functionality and architecture. \Cref{sec:installation-usage} gives instructions for the user to install the library and provides an example and a experimental framework for validating the library. \Cref{sec:quality-standards} describes the quality standards of the code developing process. \Cref{sec:conclusions} summarises the conclusions of this paper and the future work. \ref{appendix:neural-network-architectures} includes the detailed description of the networks included in the framework. Finally \ref{appendix:experimental-studies} includes the extended explanation about the cases of study.

\section{Software Description}
\label{sec:software_framework}

The \texttt{TSFE$_{DL}$} library is built on Python 3. The library follows the programming style of the \texttt{Tensorflow+Keras} \cite{keras} functional API for further integration into state-of-the-art Machine Learning pipelines and seamlessly modification of the provided models. Moreover, a \texttt{PyTorch} \cite{pytorch} implementation of the library that relies on \texttt{PyTorch-Lightning} is provided in order to allow the user both to easily scale-up model execution in multi-GPU clusters and to create new architectures maximising code re-utilisation.

\begin{figure}[!hbt]
    \centering
    \resizebox{.8\linewidth}{!}{
        \begin{tikzpicture}
          \node[input](input){\rotatebox{90}{Input}};
          \node[conv,minimum height=6cm,right=1.3cm of input](conv1){\rotatebox{90}{Conv1}};
          \node[conv,minimum height=4.5cm,right=0.2cm of conv1](conv3){\rotatebox{90}{Conv2}};
          \node[right=0.6cm of conv3](dots1){$\dots$};
          \node[conv,minimum height=2.5cm,right=0.2cm of dots1](convN){\rotatebox{90}{ConvN}};
          \node[block,right=1.5cm of convN](rnn1){RNN1};
          \node[right=0.7cm of rnn1](dots2){$\dots$};
          \node[block,right=0.3cm of dots2](rnn2){RNN2};
          \node[dense,minimum height=6cm,right=1.3cm of rnn2](dense1){\rotatebox{90}{Dense1}};
          \node[dense,minimum height=4.5cm,right=0.2cm of dense1](dense2){\rotatebox{90}{Dense2}};
          \node[right=0.6cm of dense2](dots2){$\dots$};
          \node[dense,minimum height=2.5cm,right=0.2cm of dots2](denseN){\rotatebox{90}{DenseN}};
          \node[circle,draw,radius=1cm,right=2cm of denseN](out){Output};
          
          \coordinate[below=1cm of conv1]  (x1);
          \coordinate[below=3.5cm of rnn2.south east]  (x2);
          \coordinate[below=0.9cm of dense1]  (x3);
          \coordinate[below=2.6cm of denseN.south east]  (x4);
          
          \draw[cbrace] (x1) -- (x2) node [black,midway,yshift=-0.8cm] 
        {\textbf{Spatio-temporal embedding module}};
          \draw[cbrace] (x3) -- (x4) node [black,midway,yshift=-0.8cm] 
        {\textbf{Top module}};
        
          \draw [-triangle 60,link] ([xshift=0.2cm,yshift=0.2cm]convN.east) -- ([yshift=0.2cm]rnn1.west);
          \draw [-triangle 60,link] ([xshift=0.2cm]input.east) -- (conv1.west);
          \draw [-triangle 60,link] ([xshift=0.2cm]rnn2.east) -- (dense1.west);
          \draw [-triangle 60,link] ([xshift=0.2cm]denseN.east) -- (out.west);
        \end{tikzpicture}
     }
    \caption{General scheme of the models presented in the \texttt{TSFE$_{DL}$} library.}
    \label{fig:models_scheme}
\end{figure}
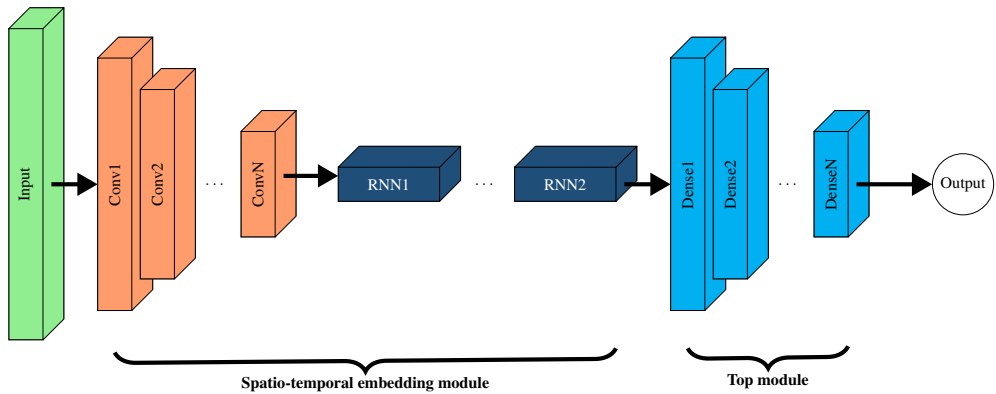

As can be seen in \Cref{fig:models_scheme}, the general architecture of the networks presented in this library is divided into two parts: the spatio-temporal embedding and the specialisation module. The former is in charge of extracting the most relevant spatio-temporal features and their bonds, whereas the specialisation module performs user-specific operations with the extracted features. This architecture allows the user to easily apply these networks to different data mining tasks just by the modification of the specialisation module. Similarly, the majority of spatio-temporal embeddings are composed of a set of CNN layers followed by a set of RNN layers such as LSTMs or GRUs which enables both the extraction of high-level spatio-temporal features together with their temporal dependencies. Additional details about the layers composition of each network are depicted in \Cref{tab:models}.

The Keras models are implemented as Python functions. Each model has an assigned function which builds it and returns a Keras \texttt{Model} object. These functions contain several parameters to customise the architecture, returning either a model with no output layers configured or a classification group of layers with a given number of classes. After the model is returned from the function the practitioner can easily add more layers using the functional programming style of Keras.

The models implemented in Pytorch are structured as classes, inheriting from the template class \texttt{TSFEDL\_BaseModule}. The template class implements the base methods to compose a PyTorch trainable network whereas each of the particular classes include the specific parameters to customise the module. These parameters include the \texttt{top\_module} parameter, which is a PyTorch \texttt{nn.Module} object for the specialisation layers for the output of the model.

The main goal of this library is to provide an easy source of models and code for the practitioners to include in their Machine Learning pipelines, gathering state-of-the art proposals for time series feature extraction. In order to achieve this, the models are programmed as reusable and customisable as possible, allowing the user to fully configure the output layers of the model to fit any type of problem such as classification, regression or forecasting. This feature is materialised by means of the \texttt{input\_shape} and \texttt{top\_module} parameters which control the input tensor shape and the output layers of the network.
 
\begin{table}[!hbt]
    \centering
    \resizebox{\linewidth}{!}{
    \tiny
    \begin{tabular}{lccccc}
        \hline
        Model names & CNN & LSTM& GRU & Bid. LSTM & Bid. GRU \\
        \hline
        CaiWenjuan \cite{CaiWenjuan} & \textcolor{green}{\Checkmark} & \textcolor{red}{\xmark} & \textcolor{red}{\xmark} & \textcolor{red}{\xmark} & \textcolor{red}{\xmark} \\
        ChenChen \cite{ChenChen} & \textcolor{green}{\Checkmark} & \textcolor{green}{\Checkmark} & \textcolor{red}{\xmark} & \textcolor{red}{\xmark} & \textcolor{red}{\xmark} \\
        FuJiangmeng \cite{FuJiangmeng} & \textcolor{green}{\Checkmark} & \textcolor{green}{\Checkmark} & \textcolor{red}{\xmark} & \textcolor{red}{\xmark} & \textcolor{red}{\xmark} \\
        GaoJunli \cite{GaoJunli} & \textcolor{red}{\xmark} & \textcolor{green}{\Checkmark} & \textcolor{red}{\xmark} & \textcolor{red}{\xmark} & \textcolor{red}{\xmark} \\
        GenMinxing \cite{GenMinxing} & \textcolor{red}{\xmark} & \textcolor{red}{\xmark} & \textcolor{red}{\xmark} & \textcolor{green}{\Checkmark} & \textcolor{red}{\xmark} \\
        HongTan \cite{HongTan} & \textcolor{green}{\Checkmark} & \textcolor{green}{\Checkmark} & \textcolor{red}{\xmark} & \textcolor{red}{\xmark} & \textcolor{red}{\xmark} \\
        HtetMyetLynn \cite{HtetMyetLynn} & \textcolor{green}{\Checkmark} & \textcolor{red}{\xmark} & \textcolor{red}{\xmark} & \textcolor{green}{\Checkmark} & \textcolor{green}{\Checkmark} \\
        HuangMeiLing \cite{HuangMeiLing} & \textcolor{green}{\Checkmark} & \textcolor{red}{\xmark} & \textcolor{red}{\xmark} & \textcolor{red}{\xmark} & \textcolor{red}{\xmark} \\
        KhanZulfiqar \cite{KhanZulfiqar} & \textcolor{green}{\Checkmark} & \textcolor{red}{\xmark} & \textcolor{green}{\Checkmark} & \textcolor{red}{\xmark} & \textcolor{red}{\xmark} \\
        KimTaeYoung \cite{KimTaeYoung} & \textcolor{green}{\Checkmark} & \textcolor{green}{\Checkmark} & \textcolor{red}{\xmark} & \textcolor{red}{\xmark} & \textcolor{red}{\xmark} \\
        KongZhengmin \cite{KongZhengmin} & \textcolor{green}{\Checkmark} & \textcolor{green}{\Checkmark} & \textcolor{red}{\xmark} & \textcolor{red}{\xmark} & \textcolor{red}{\xmark} \\
        LihOhShu \cite{LihOhShu} & \textcolor{green}{\Checkmark} & \textcolor{green}{\Checkmark} & \textcolor{red}{\xmark} & \textcolor{red}{\xmark} & \textcolor{red}{\xmark} \\
        OhShuLih \cite{OhShuLih} & \textcolor{green}{\Checkmark} & \textcolor{green}{\Checkmark} & \textcolor{red}{\xmark} & \textcolor{red}{\xmark} & \textcolor{red}{\xmark} \\
        ShiHaotian \cite{ShiHaotian} & \textcolor{green}{\Checkmark} & \textcolor{green}{\Checkmark} & \textcolor{red}{\xmark} & \textcolor{red}{\xmark} & \textcolor{red}{\xmark} \\
        WangKejun \cite{WangKejun} & \textcolor{green}{\Checkmark} & \textcolor{green}{\Checkmark} & \textcolor{red}{\xmark} & \textcolor{red}{\xmark} & \textcolor{red}{\xmark} \\
        WeiXiaoyan \cite{WeiXiaoyan} & \textcolor{green}{\Checkmark} & \textcolor{green}{\Checkmark} & \textcolor{red}{\xmark} & \textcolor{red}{\xmark} & \textcolor{red}{\xmark} \\
        YaoQihang \cite{YaoQihang} & \textcolor{green}{\Checkmark} & \textcolor{green}{\Checkmark} & \textcolor{red}{\xmark} & \textcolor{red}{\xmark} & \textcolor{red}{\xmark} \\
        YiboGao \cite{YiboGao} & \textcolor{green}{\Checkmark} & \textcolor{red}{\xmark} & \textcolor{red}{\xmark} & \textcolor{red}{\xmark} & \textcolor{red}{\xmark} \\
        YildirimOzal \cite{YildirimOzal} & \textcolor{green}{\Checkmark} & \textcolor{green}{\Checkmark} & \textcolor{red}{\xmark} & \textcolor{red}{\xmark} & \textcolor{red}{\xmark} \\
        ZhangJin \cite{ZhangJin} & \textcolor{green}{\Checkmark} & \textcolor{red}{\xmark} & \textcolor{red}{\xmark} & \textcolor{red}{\xmark} & \textcolor{green}{\Checkmark} \\
        ZhengZhenyu \cite{ZhengZhenyu} & \textcolor{green}{\Checkmark} & \textcolor{green}{\Checkmark} & \textcolor{red}{\xmark} & \textcolor{red}{\xmark} & \textcolor{red}{\xmark} \\
        \hline
    \end{tabular}
    }
    \caption{TSFE$_{DL}$ models and architecture type.}
    \label{tab:models}
\end{table}

\section{Installation and Quality Standards}
\label{sec:installation-usage}

The \texttt{TSFE$_{DL}$} library can be installed using PyPi through the command \texttt{pip install TSFEDL}. It is also available by cloning the repository from GitHub\footnote{https://github.com/ari-dasci/S-TSFE-DL} and executing, from the root directory, the command \texttt{python setup.py install}. After that, the package will be available for its usage within the name \texttt{TSFEDL}.

The library code follows the \texttt{PEP8} style standard for Python. \texttt{Travis-CI} service is enabled in the repository of the project for continuous integration, ensuring back-compatibility and a proper operation of the architectures. \textit{Semantic Versioning} and \textit{Keep a Changelog} standards are integrated in the repository as well, making it easier for the users to notice the changes in the version progression. An extensive documentation is provided, following the \texttt{numpydoc} style of comments and using \texttt{sphinx} to generate it. The documentation is hosted in the \texttt{Read the Docs}\footnote{https://s-tsfe-dl.readthedocs.io/en/latest/} platform.

\section{Cases of Study}
\label{sec:quality-standards}

An experimental framework is proposed to study the performance of the networks and the capabilities of the library. In this section, a new model leveraging the characteristics of \texttt{TSFE$_{DL}$} is created, remarking the customisation capability for any given task. The objective with the custom model is to extract new spatio-temporal features to outperform the included architectures in the library on several kinds of problems. The problems analysed are briefly described below: 

\begin{enumerate}
    \item Forecasting. The aim is to predict the following $n=50$ timesteps of a time series from the Spanish Digital Seismic Network \cite{IGN} to identify future earthquakes \footnote{IGN Data under contact https://www.ign.es/web/ign/portal/sis-area-sismicidad}.
    
    \item Classification. The objective is to classify segments of ECG signals from the MIT-BIH dataset \cite{MIT_BIH} to identify different types of cardiac arrhythmia \footnote{MIT-BIH data https://physionet.org/content/mitdb/1.0.0/}.
    
    \item Anomaly Detection. The goal is to identify malicious attacks from network traffic using the KDD Cup '99 dataset \cite{kddcup99} \footnote{KDDCup99 data http://kdd.ics.uci.edu/databases/kddcup99/kddcup99.html}.
\end{enumerate}

The network shown in \Cref{alg:example} is a modification of the \texttt{HuangMeiLing} model which includes a new LSTM layer. Next, the spatio-temporal features extracted by this model will be further processed using a specialisation module to fulfil the requirements of each task. In \Cref{alg:example}, an example for the forecasting task is shown. Additional details about these experiments and its characteristics can be found in the library repository\footnote{\url{https://github.com/ari-dasci/S-TSFE-DL/tree/main/examples}}. We must highlight that this specialisation module is employed on all the networks of the library following the same procedure as shown in \Cref{alg:example} to perform a fair comparison between methods.

\begin{figure}[!htbp]
    \begin{Verbatim}[commandchars=\\\{\}]
    \PYG{k+kn}{import} \PYG{n+nn}{tensorflow} \PYG{k}{as} \PYG{n+nn}{tf}
    \PYG{k+kn}{import} \PYG{n+nn}{TSFEDL.models\PYGZus{}keras} \PYG{k}{as} \PYG{n+nn}{TSFEDL}
    
    \PYG{n+nb}{input} \PYG{o}{=} \PYG{n}{tf}\PYG{o}{.}\PYG{n}{keras}\PYG{o}{.}\PYG{n}{Input}\PYG{p}{(}\PYG{n}{shape}\PYG{o}{=}\PYG{p}{(}\PYG{l+m+mi}{1000}\PYG{p}{,} \PYG{l+m+mi}{1}\PYG{p}{))}
    \PYG{n}{model} \PYG{o}{=} \PYG{n}{TSFEDL}\PYG{o}{.}\PYG{n}{HuangMeiLing}\PYG{p}{(}\PYG{n}{input\PYGZus{}tensor}\PYG{o}{=}\PYG{n+nb}{input}\PYG{p}{,} \PYG{n}{include\PYGZus{}top}\PYG{o}{=}\PYG{k+kc}{False}\PYG{p}{)}
    \PYG{n}{x} \PYG{o}{=} \PYG{n}{model}\PYG{o}{.}\PYG{n}{output}
    \PYG{n}{x} \PYG{o}{=} \PYG{n}{tf}\PYG{o}{.}\PYG{n}{keras}\PYG{o}{.}\PYG{n}{layers}\PYG{o}{.}\PYG{n}{LSTM}\PYG{p}{(}\PYG{n}{units}\PYG{o}{=}\PYG{l+m+mi}{20}\PYG{p}{)(}\PYG{n}{x}\PYG{p}{)}
    \PYG{c+c1}{\PYGZsh{} Add the top module}
    \PYG{n}{x} \PYG{o}{=} \PYG{n}{tf}\PYG{o}{.}\PYG{n}{keras}\PYG{o}{.}\PYG{n}{layers}\PYG{o}{.}\PYG{n}{Flatten}\PYG{p}{(}\PYG{n}{x}\PYG{p}{)}
    \PYG{n}{x} \PYG{o}{=} \PYG{n}{tf}\PYG{o}{.}\PYG{n}{keras}\PYG{o}{.}\PYG{n}{layers}\PYG{o}{.}\PYG{n}{Dense}\PYG{p}{(}\PYG{l+m+mi}{50}\PYG{p}{)(}\PYG{n}{x}\PYG{p}{)}
    \PYG{n}{out} \PYG{o}{=} \PYG{n}{tf}\PYG{o}{.}\PYG{n}{keras}\PYG{o}{.}\PYG{n}{layers}\PYG{o}{.}\PYG{n}{Reshape}\PYG{p}{([}\PYG{l+m+mi}{50}\PYG{p}{,}\PYG{l+m+mi}{1}\PYG{p}{])(}\PYG{n}{x}\PYG{p}{)}
    \PYG{c+c1}{\PYGZsh{} Create the new model and train it.}
    \PYG{n}{new\PYGZus{}model} \PYG{o}{=} \PYG{n}{tf}\PYG{o}{.}\PYG{n}{keras}\PYG{o}{.}\PYG{n}{Model}\PYG{p}{(}\PYG{n}{inputs}\PYG{o}{=}\PYG{n+nb}{input}\PYG{p}{,} \PYG{n}{outputs}\PYG{o}{=}\PYG{n}{out}\PYG{p}{)}
    \PYG{n}{new\PYGZus{}model}\PYG{o}{.}\PYG{n}{compile}\PYG{p}{(}\PYG{n}{loss}\PYG{o}{=}\PYG{l+s+s1}{\PYGZsq{}mae\PYGZsq{}}\PYG{p}{,} \PYG{n}{optimizer}\PYG{o}{=}\PYG{l+s+s1}{\PYGZsq{}adam\PYGZsq{}}\PYG{p}{,} \PYG{n}{metrics}\PYG{o}{=}\PYG{p}{[}\PYG{l+s+s1}{\PYGZsq{}mae\PYGZsq{}}\PYG{p}{])}
    \PYG{n}{new\PYGZus{}model}\PYG{o}{.}\PYG{n}{fit}\PYG{p}{(}\PYG{n}{data}\PYG{p}{,} \PYG{n}{epochs}\PYG{o}{=}\PYG{l+m+mi}{20}\PYG{p}{)}
    \end{Verbatim}
    \caption{Creation of a new CNN-RNN model for time series forecasting using \texttt{TSFE$_{DL}$}.}
    \label{alg:example}
\end{figure}

The results obtained from each method on each task are shown in \Cref{tab:resultados}. These results led us to the conclusion that the models in this library can be applied to different tasks successfully, even if they have not been initially designed for them. In fact, no method outperforms all the rest for all of the problems. The custom model created in \Cref{alg:example} outperforms the rest of the networks in the forecasting task, confirming that the customisation capability can enhance the performance of the networks. This means that the library's variety of models, together with its customisation, gives great flexibility concerning the target application.

\begin{table}[!hbtp]
    \caption{Results extracted from the \texttt{TSFE$_{DL}$} models for the three tasks analysed.}
    \label{tab:resultados}
    \resizebox{\linewidth}{!}{
    \scriptsize
    \begin{tabular}{lrrr}
        \hline
        \textbf{Model} & \multicolumn{1}{l}{\textbf{\begin{tabular}[c]{@{}l@{}}Classification \\ (Accuracy) \\ MIT-BIH arrythmia\end{tabular}}} & \multicolumn{1}{l}{\textbf{\begin{tabular}[c]{@{}l@{}}Time series forecasting \\ (MAE) \\ Spanish Digital Seismic Network\end{tabular}}} & \multicolumn{1}{l}{\textbf{\begin{tabular}[c]{@{}l@{}}Anomaly detection \\ (AUC) \\ KDD Cup99\end{tabular}}} \\ \hline
        CaiWenjuan \cite{CaiWenjuan}     & 0.6845 & 3.8784 & 0.5945 \\
        ChenChen \cite{ChenChen}       & 0.9233 & 151.3446 & \textbf{0.7402} \\
        FuJiangmeng \cite{FuJiangmeng}   & 0.5612 & 12.0492 & 0.5135 \\
        GaoJunLi \cite{GaoJunli}    & 0.4996 & 45.9238 & 0.4821 \\
        GenMinxing  \cite{GenMinxing}   & 0.9569 & 6.8440 & 0.5186 \\
        HongTan \cite{HongTan}   & 0.8412 & 9.1740 & 0.6069 \\ 
        HtetMyetLynn \cite{HtetMyetLynn}  & 0.9709 & 12.1476 & 0.4866 \\
        HuangMeiLing \cite{HuangMeiLing} & 0.9633 & 1.4135 & 0.5180 \\
        KhanZulfiqar \cite{KhanZulfiqar}  & 0.9400 & 29.9031 & 0.6272 \\
        KimTaeYoung \cite{KimTaeYoung}  & 0.6378 & 3.1309 & 0.5586 \\
        KongZhengmin \cite{KongZhengmin}  & 0.6515 & 2.0443 & 0.5535 \\
        LihOhShu  \cite{LihOhShu}     & 0.8196 & 3.0234 & 0.5302 \\
        OhShuLih  \cite{OhShuLih}       & 0.7224 & 2.2590 & 0.7103 \\
        ShiHaotian \cite{ShiHaotian}    & 0.9581 & 1.9449 & 0.6071 \\
        WangKejun   \cite{WangKejun}   & 0.9539 & 3.0436 & 0.4897 \\
        WeiXiaoyan  \cite{WeiXiaoyan}   & 0.9703 & 2.9720 & 0.6462 \\
        YaoQihang  \cite{YaoQihang}    & 0.9681 & 2.7898 & 0.5939 \\
        YiboGao \cite{YiboGao}       & \textbf{0.9718} & 149.2162 & 0.7356 \\
        YildirimOzal \cite{YildirimOzal}  & 0.9075 & 44.5272 & 0.6988  \\
        ZhangJin    \cite{ZhangJin}   & 0.9575 & 30.6660 & 0.5605 \\
        ZhengZhenyu \cite{ZhengZhenyu}   & 0.9196 & 2.0847 & 0.5374 \\ 
        Model of the example (\Cref{alg:example}) & 0.9248 & \textbf{1.1402}  & 0.5089\\ \hline
    
    \end{tabular}
    }
\end{table}

\section{Concluding Remarks}
\label{sec:conclusions}

The Python library \texttt{TSFE$_{DL}$} gathers 20 Deep Learning state-of-the-art methods combining both convolutional and recurrent layers. The implementation relies on the \texttt{Keras} functional API and \texttt{PyTorch-Lightning} to easily integrate the algorithms into state-of-the-art Machine Learning pipelines. This fact, together with the architecture of the library, allows us to easily create and customise Deep Learning models for different data mining tasks. 

As final conclusions it may be remarked that the library provides performant, expandable and customisable neural networks being proved the usefulness of the provided piece of software. The results confirm that the included models can be successfully applied to different tasks. Therefore, the application scope of this type of models can be significantly extended. This Python module stands as a convenient solution for the practitioner.

\section*{Acknowledgements}

This work has been partially supported by the Contract UGR-AM OTRI-4260 and the Regional Government of Andalusia, under the program ``Personal Investigador Doctor'', reference DOC\_00235. This work was also supported by project PID2020-119478GB-I00 granted by Ministerio de Ciencia, Innovaci\'on y Universidades, and projects P18-FR-4961 and P18-FR-4262 by Proyectos I+D+i Junta de Andalucia 2018.

\bibliographystyle{unsrt}
\bibliography{bibliography.bib}

\appendix
\section{Neural Network Architectures}
\label{appendix:neural-network-architectures}

In this appendix the different architectures implemented will be detailed, explaining the layers architecture and the purpose of the models.

\subsection{CaiWenjuan}
\label{subsection:CaiWenjuan}

The architecture \cite{CaiWenjuan} is designed to detect atrial fibrillation from electrocardiograms. 

\begin{figure}[!hbt]
    \centering
    \includegraphics[width=\textwidth]{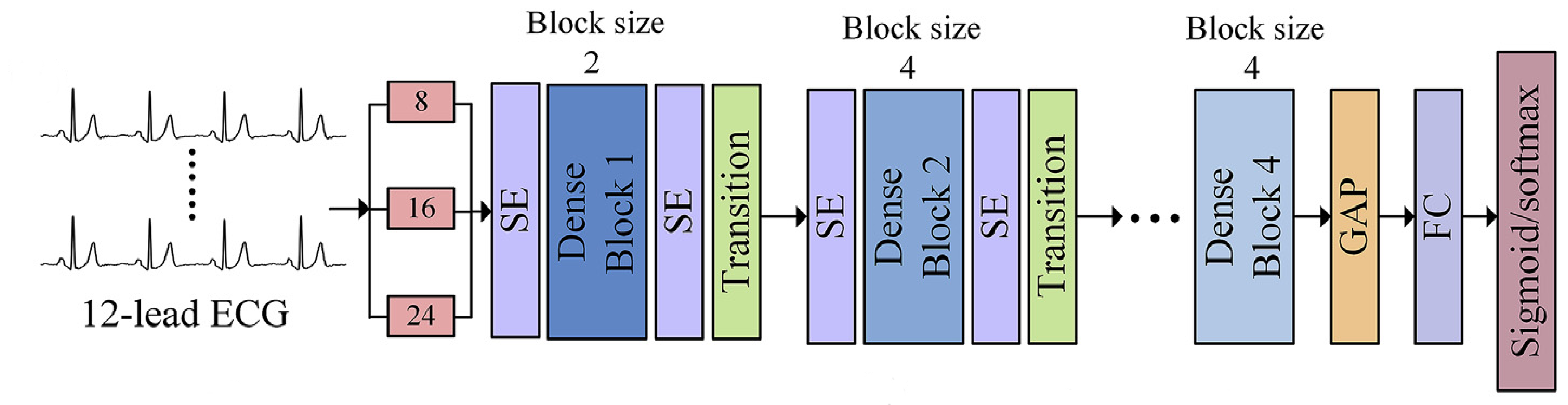}
    \caption{CaiWenjuan model (extracted from \cite{CaiWenjuan}).}
    \label{fig:CaiWenjuan}
\end{figure}

\begin{figure}[!hbt]
    \centering
    \includegraphics[scale=0.1]{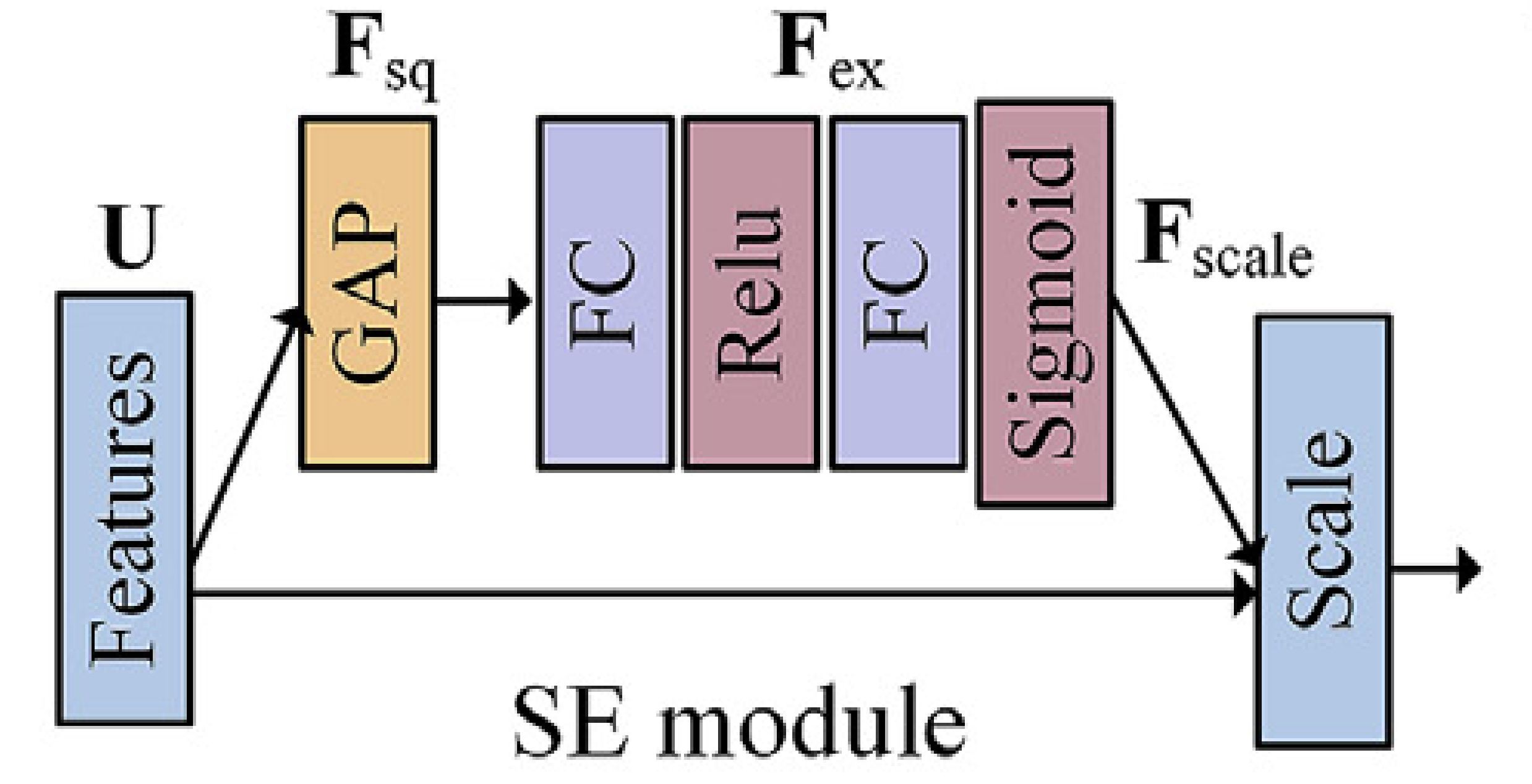}
    \caption{CaiWenjuan squeeze and excitation module (extracted from \cite{CaiWenjuan}).}
    \label{fig:CaiWenjuan-squeeze-excitation}
\end{figure}

The architecture of the network can be seen in \Cref{fig:CaiWenjuan}. In this structure GAP stands for Global Average Pooling, SE for Squeeze and Excitation block and FC for Fully Connected. The non-trivial modules and layers will be explained in this section.

The network structure starts with one-dimensional convolutions to extract the features from the time series. The input is passed through 3 non-consecutive convolutions, having the same input passed to three different convolution layers to extract different feature sizes, which are composed by concatenation. 

The squeeze and excitation module \cite{squeeze_and_activation} (\Cref{fig:CaiWenjuan-squeeze-excitation}) is used to improve the quality of the representation by using global average pooling, fully connected layers, the RELU and sigmoid activation functions to modify the signal and finally a rescale of the features.

The dense blocks have different block sizes. For each of the blocks inside the dense block a batch normalization layer, a RELU activation, a one-dimensional convolution, a batch normalization, a RELU activation function, a one-dimensional convolution and a concatenation layer can be found. The final concatenation layer defines a skip connection between the input of the block and the features extracted by the convolution pipeline described inside the dense block.

\subsection{ChenChen}
\label{subsection:ChenChen}

The ChenChen model \cite{ChenChen} is designed to detect arrythmia from electrocardiograms. 

\begin{figure}[!hbt]
    \centering
    \includegraphics[width=\textwidth]{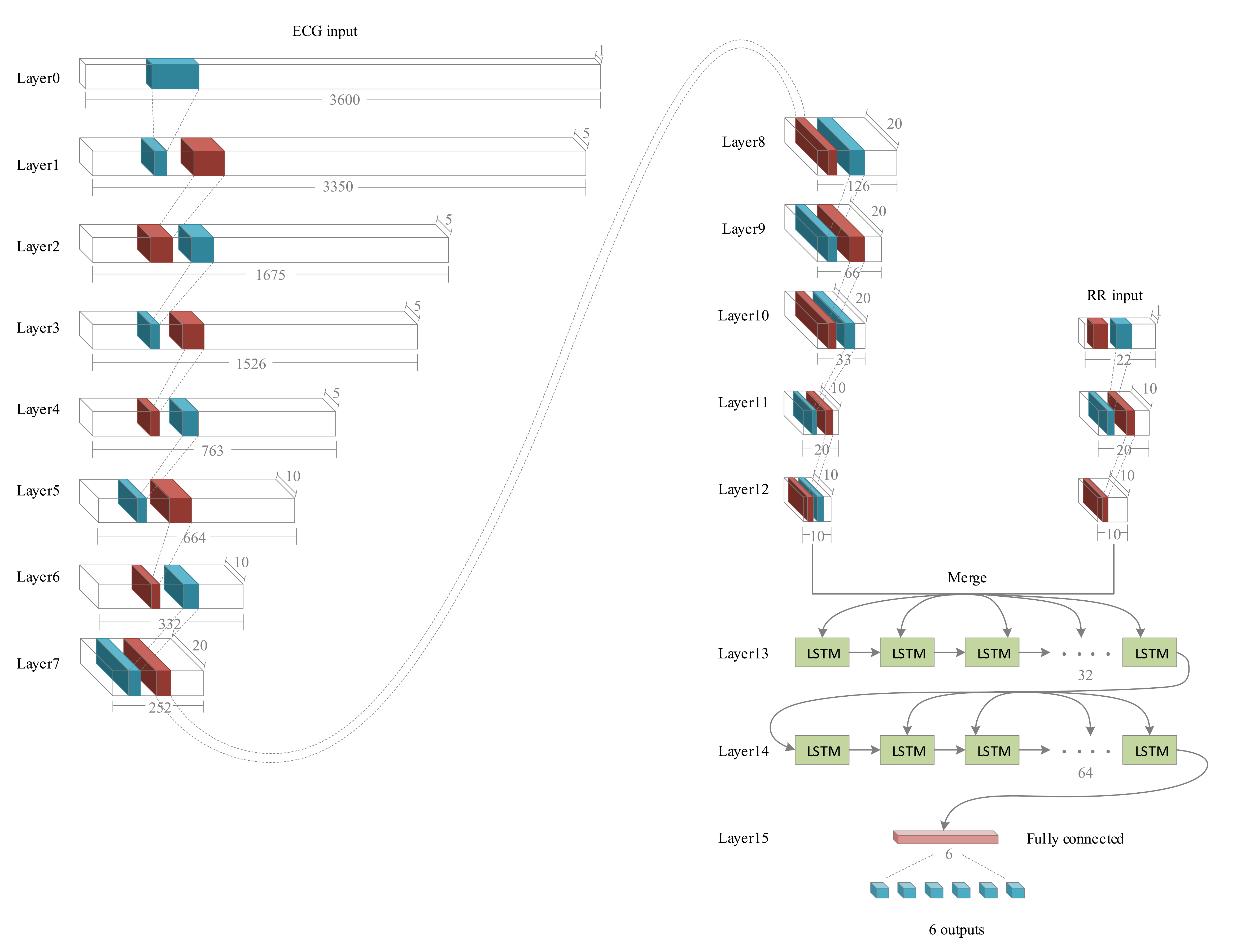}
    \caption{ChenChen model (extracted from \cite{ChenChen}).}
    \label{fig:ChenChen}
\end{figure}

The ChenChen architecture, described in \Cref{fig:ChenChen}, uses 6 one-dimensional convolution layers with 6 max pooling layers in between. This structure allows the model to extract features from the time series. After the convolutional layers two LSTM layers are used to extract the temporal dependencies of the features. The LSTM layers receive the concatenated output of the convolutions and characteristics of the electrocardiograms extracted by hand. For this library only the CNN-LSTM structure is maintained. 

\subsection{FuJiangmeng}
\label{subsection:FuJiangmeng}

The FuJiangmeng model \cite{FuJiangmeng} is designed to detect actuator failures from UAVs (Unmanned Aerial Vehicles). 

\begin{figure}[!hbt]
    \centering
    \includegraphics[width=\textwidth]{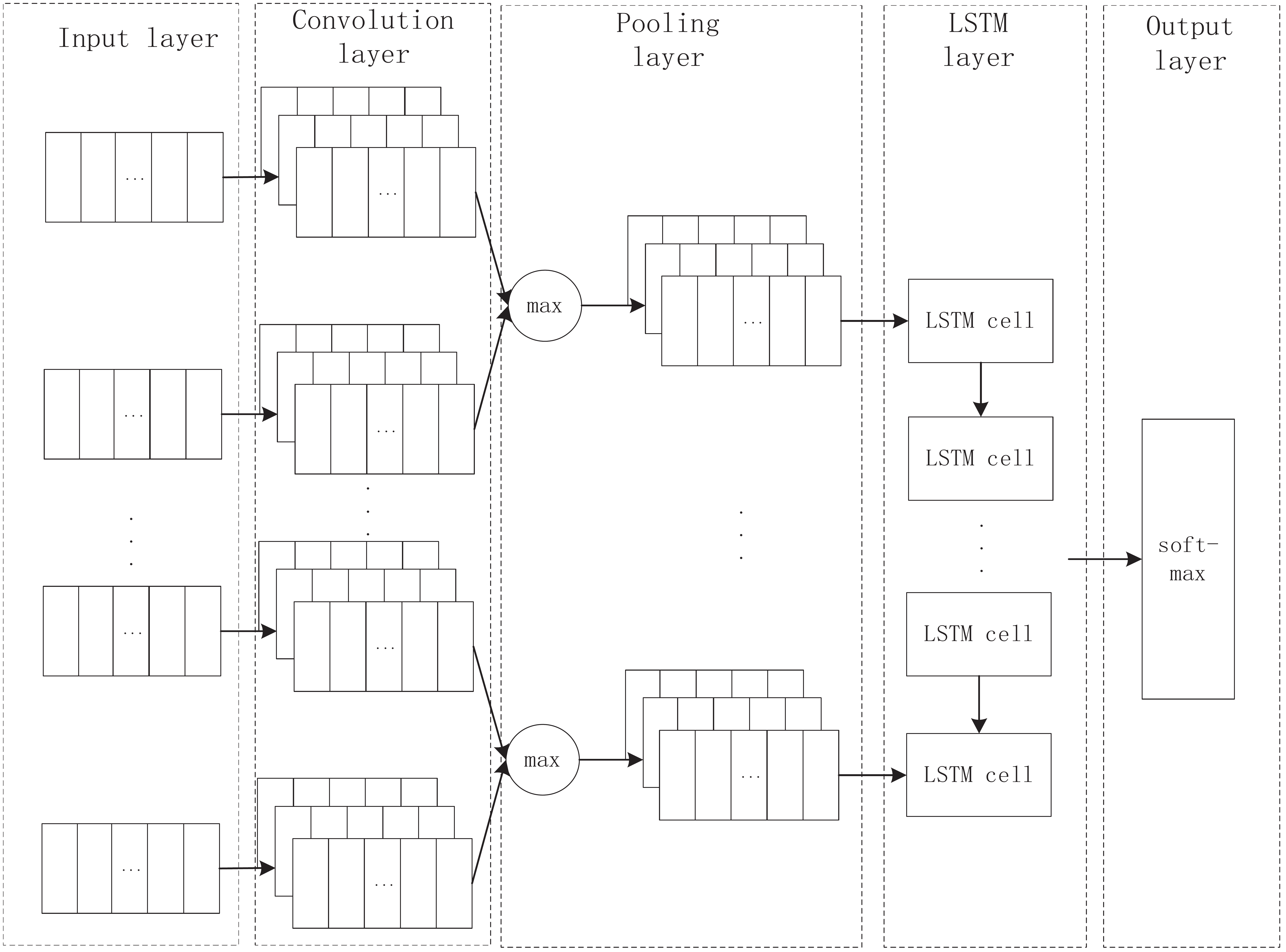}
    \caption{FuJiangmeng model (extracted from \cite{FuJiangmeng}).}
    \label{fig:FuJiangmeng}
\end{figure}

The structure of this model (\Cref{fig:FuJiangmeng}) contains one one-dimensional convolution layer, a max pooling layer and one LSTM layer. This model is a simple approach to CNN-LSTM network architecture but is a suitable alternative for smaller and simpler datasets.

\subsection{GaoJunli}
\label{subsection:GaoJunli}

The GaoJunli model \cite{GaoJunli} is designed to detect arrythmia from electrocardiograms.

\begin{figure}[!hbt]
    \centering
    \includegraphics[width=\textwidth]{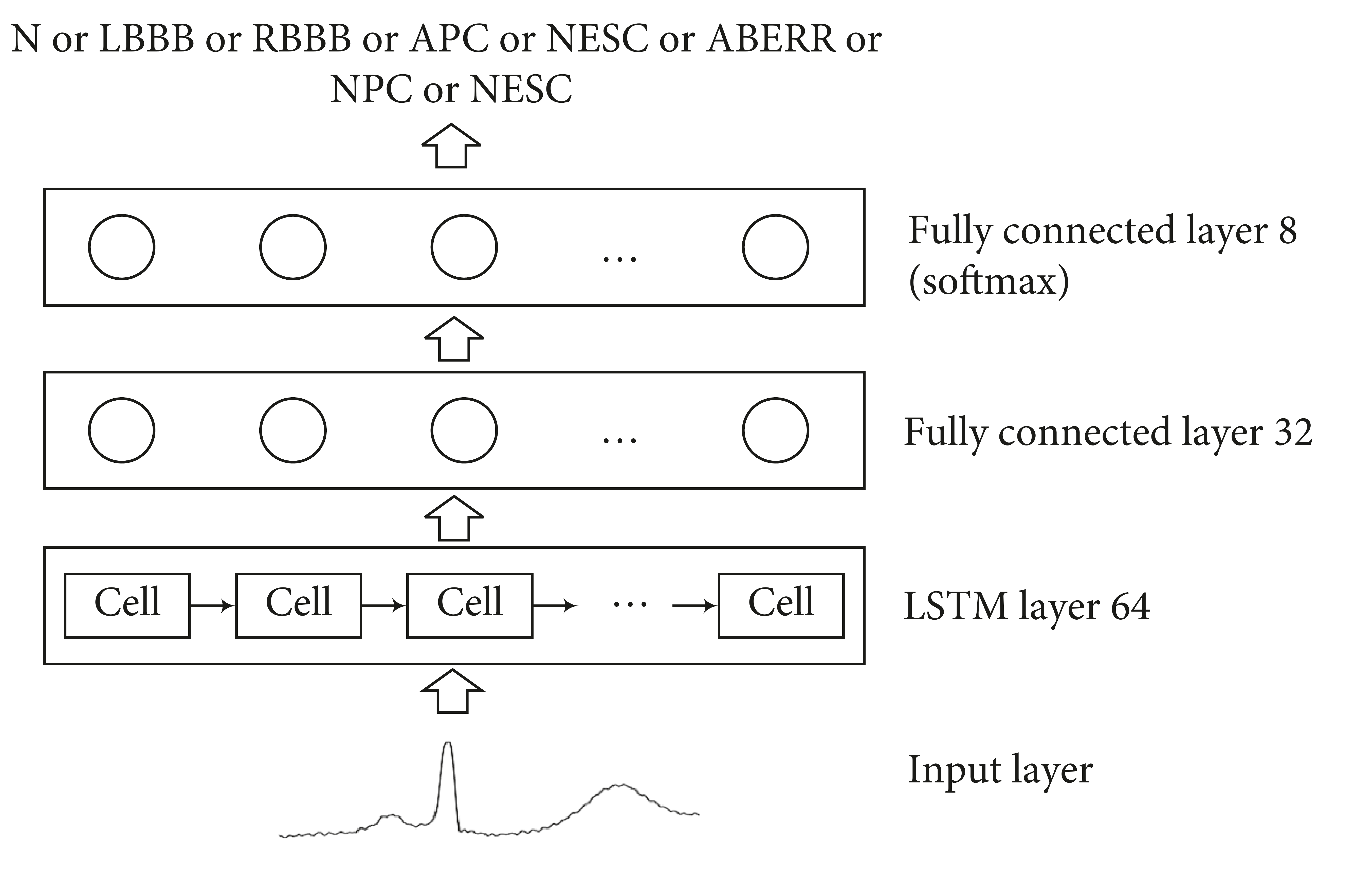}
    \caption{GaoJunli model (extracted from \cite{GaoJunli}).}
    \label{fig:GaoJunli}
\end{figure}

For this model (\Cref{fig:GaoJunli}) the authors selected manually the characteristics from the electrocardiograms. This is the reason why this model does not includes any one-dimensional convolution. The core of the model is composed only by one LSTM layer, being a simple but useful model when providing manually selected or crafted features from the data.

\subsection{GenMinxing}
\label{subsection:GenMinxing}

The GenMinxing model \cite{GenMinxing} is designed to detect epileptic seizure from electroencephalograms.

\begin{figure}[!hbt]
    \centering
    \includegraphics[width=\textwidth]{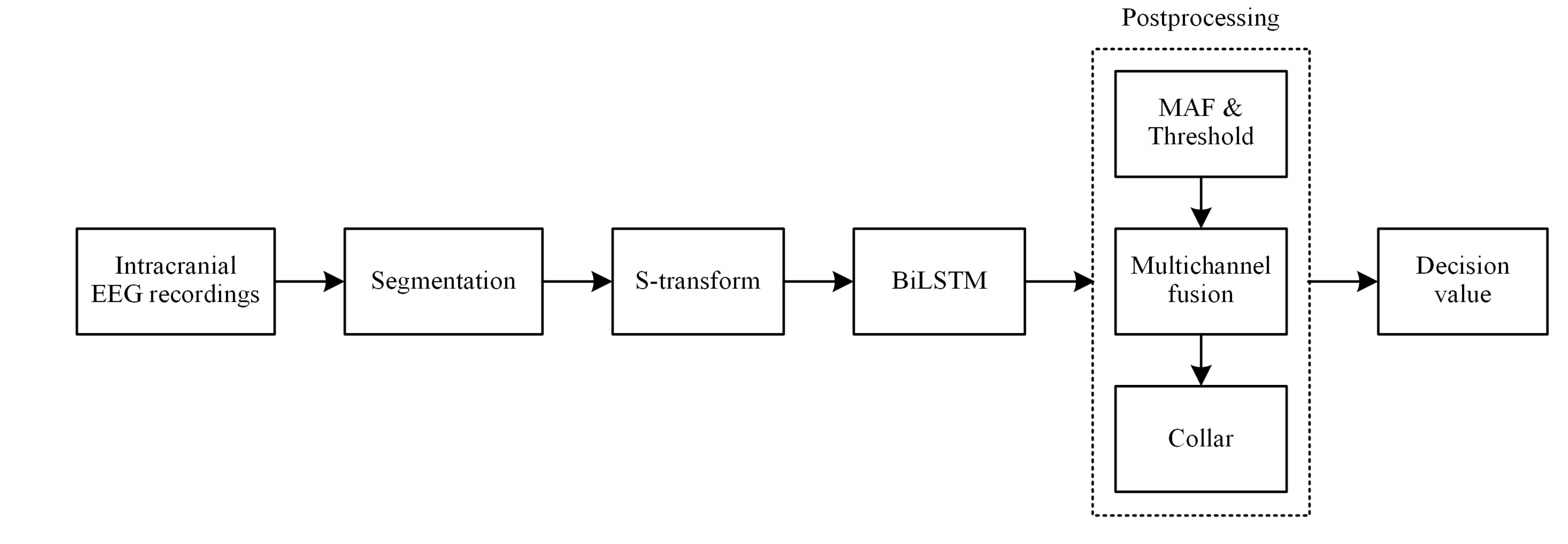}
    \caption{GenMinxing model (extracted from \cite{GenMinxing}).}
    \label{fig:GenMinxing}
\end{figure}

This neural network model (\Cref{fig:GenMinxing}) receives as input the features extracted from electroencephalograms by segmenting the interesting events and applying a time-frequency decomposition using the Stockwell transform, which combines the Fourier and the wavelet transforms. This model is only composed by one bidirectional LSTM layer. The bidirectional LSTM is a layer that uses two stacked LSTM layers which read the data in both directions (left to right and right to left). By doing so the layer extract patterns in both directions to better model the seen data and its context.

\subsection{HtetMyetLynn}
\label{subsection:HtetMyetLynn}

The HtetMyetLynn model \cite{HtetMyetLynn} is designed to detect arrhythmia from electrocardiograms.

\begin{figure}[!hbt]
    \centering
    \includegraphics[scale=0.05]{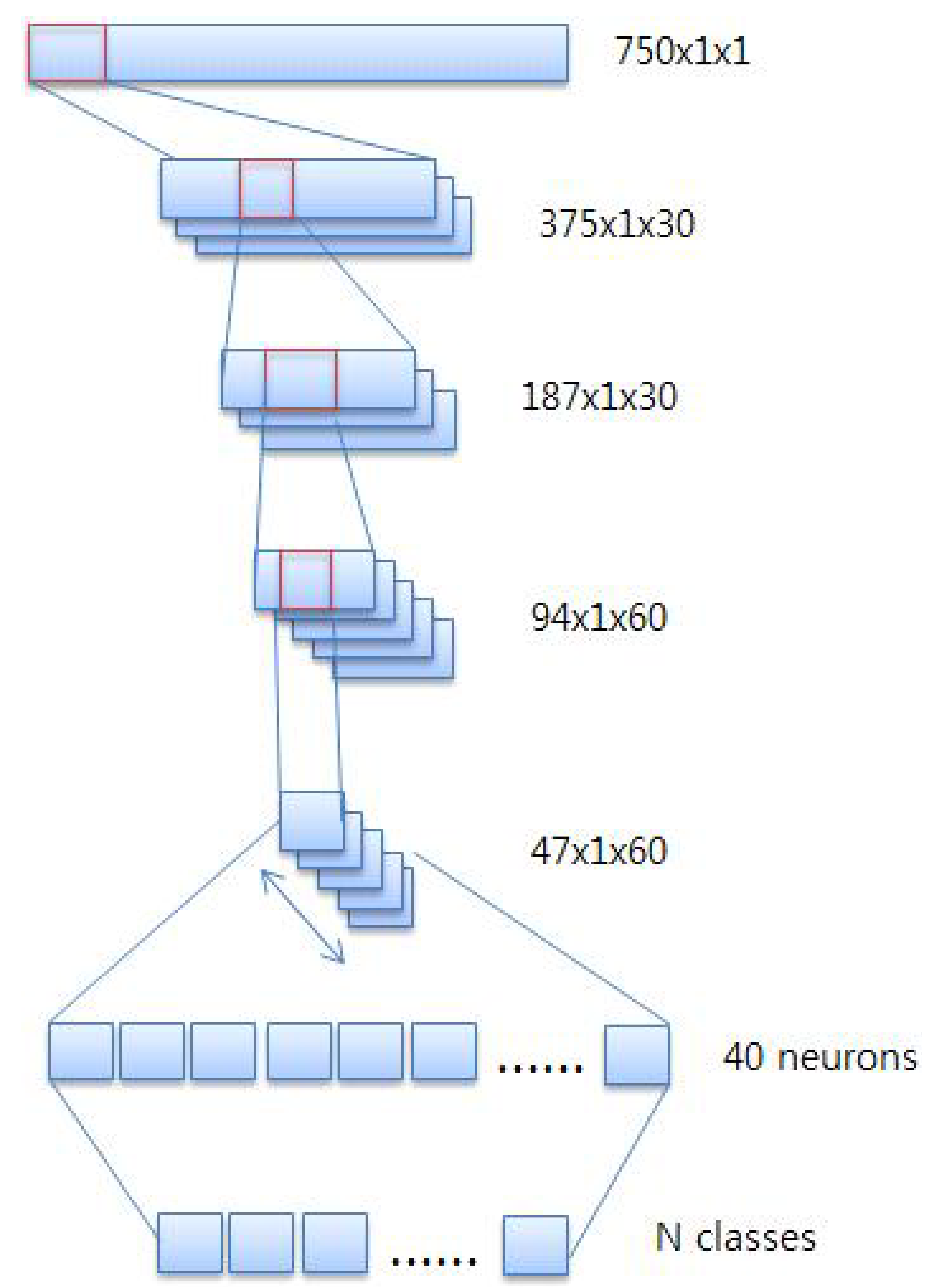}
    \caption{HtetMyetLynn model (extracted from \cite{HtetMyetLynn}).}
    \label{fig:HtetMyetLynn}
\end{figure}

The model presented in \Cref{fig:HtetMyetLynn} is a hybrid network combining one-dimensional convolutions with bidirectional recurrent layers. In the first section we can find four one-dimensional convolutional layers which extract features from the time series. These features are then passed to one bidirectional recurrent layer, which can be either LSTM or GRU. The bidirectional layer enables the model to extract temporal dependencies in both directions of the time series. LSTM layers include the memory cells which are not present in GRU, nevertheless GRU can be used to gain efficiency and when we don't face gradient explosion with our dataset.

\subsection{HuangMeiLing}
\label{subsection:HuangMeiLing}

The HuangMeiLing model \cite{HuangMeiLing} is designed to detect atrial fibrillation and normal sinus rhythm from electrocardiograms.

\begin{figure}[!hbt]
    \centering
    \includegraphics[width=\textwidth]{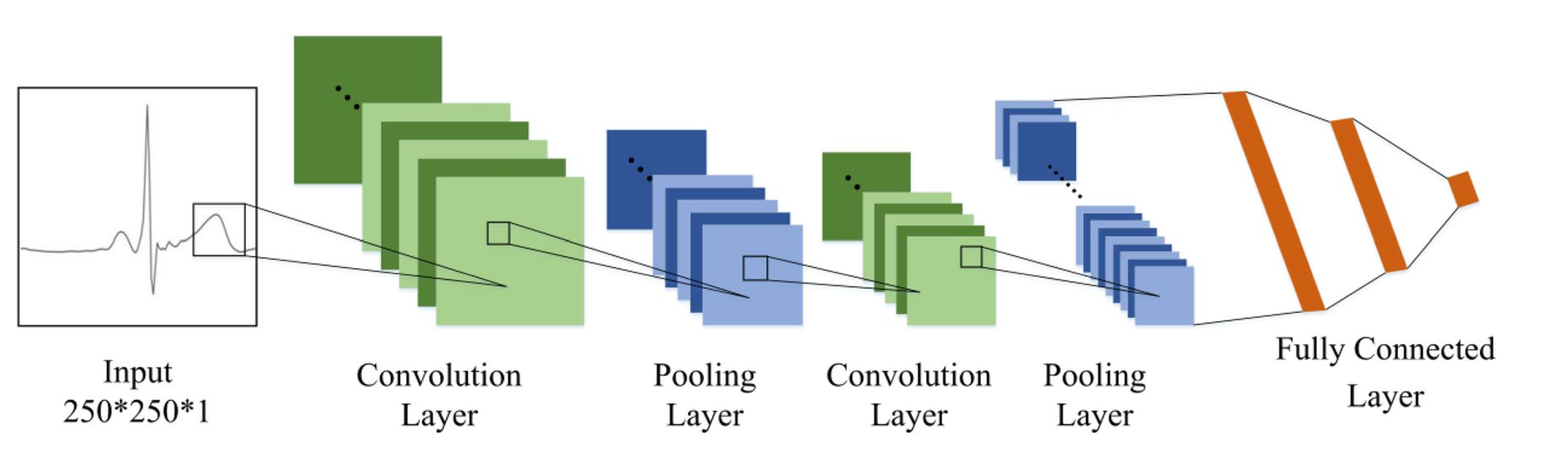}
    \caption{HuangMeiLing model (extracted from \cite{HuangMeiLing}).}
    \label{fig:HuangMeiLing}
\end{figure}

The model presented in \Cref{fig:HuangMeiLing} is a convolutional model with no recurrent layers. This architecture was originally designed to work with 2D images from the time series, but has been adapted to one-dimensional convolutions to work with most of the data. The model consists on two blocks of one-dimensional convolutions and max pooling layer. This architecture is composed only by convolutional layers to keep the structure of the network as simple as possible, therefore not extracting temporal dependencies. This model can be useful when no temporal dependencies are found in our problem or when the detection of the events can be performed obtaining only the spatial features, saving a significant amount of computation time.

\subsection{KhanZulfiqar}
\label{subsection:KhanZulfiqar}

The KhanZulfiqar model \cite{KhanZulfiqar} is designed to predict the electricity consumption in residential areas.

\begin{figure}[!hbt]
    \centering
    \includegraphics[width=\textwidth]{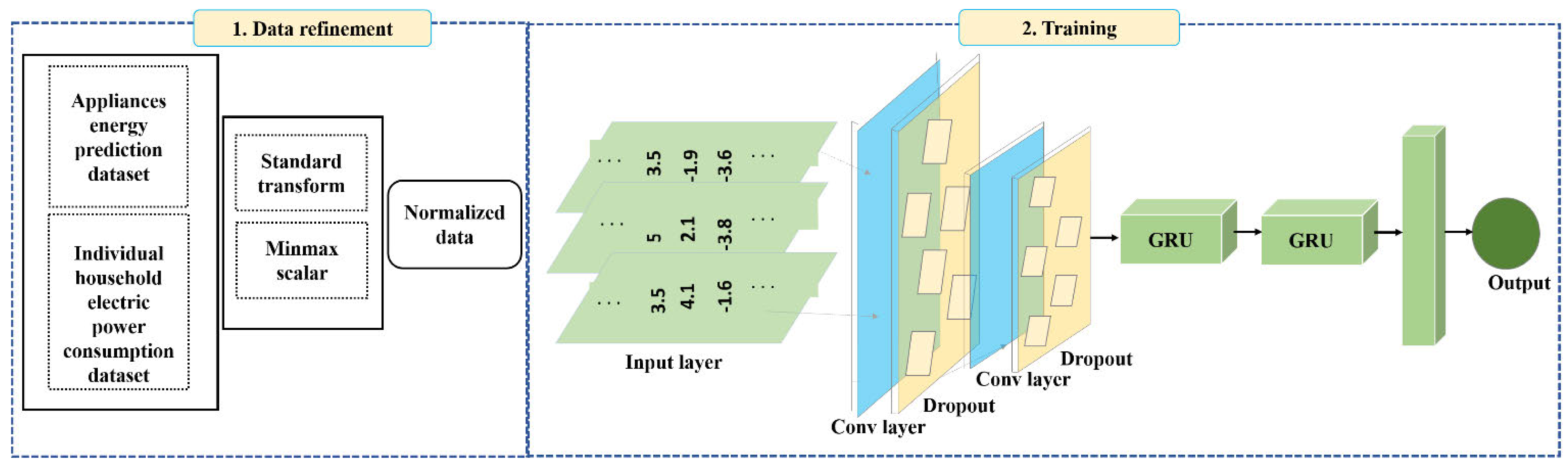}
    \caption{KhanZulfiqar model (extracted from \cite{KhanZulfiqar}).}
    \label{fig:KhanZulfiqar}
\end{figure}

The model described in \Cref{fig:KhanZulfiqar} is a hybrid model composed by one-dimensional convolutions and GRU recurrent layers. The architecture starts with two convolutional layers followed by dropout layers in order to build a better model for generalisation. After this, two GRU layers are applied in order to extract temporal dependencies. This model uses GRU over LSTM to make a faster and more efficient model, as the preprocessing of the data prevents situations where the gradient of the neural network could explode.

\subsection{KimTaeYoung}
\label{subsection:KimTaeYoung}

The KimTaeYoung model \cite{KimTaeYoung} is designed to predict the electricity consumption in residential areas.

\begin{figure}[!hbt]
    \centering
    \includegraphics[scale=0.05]{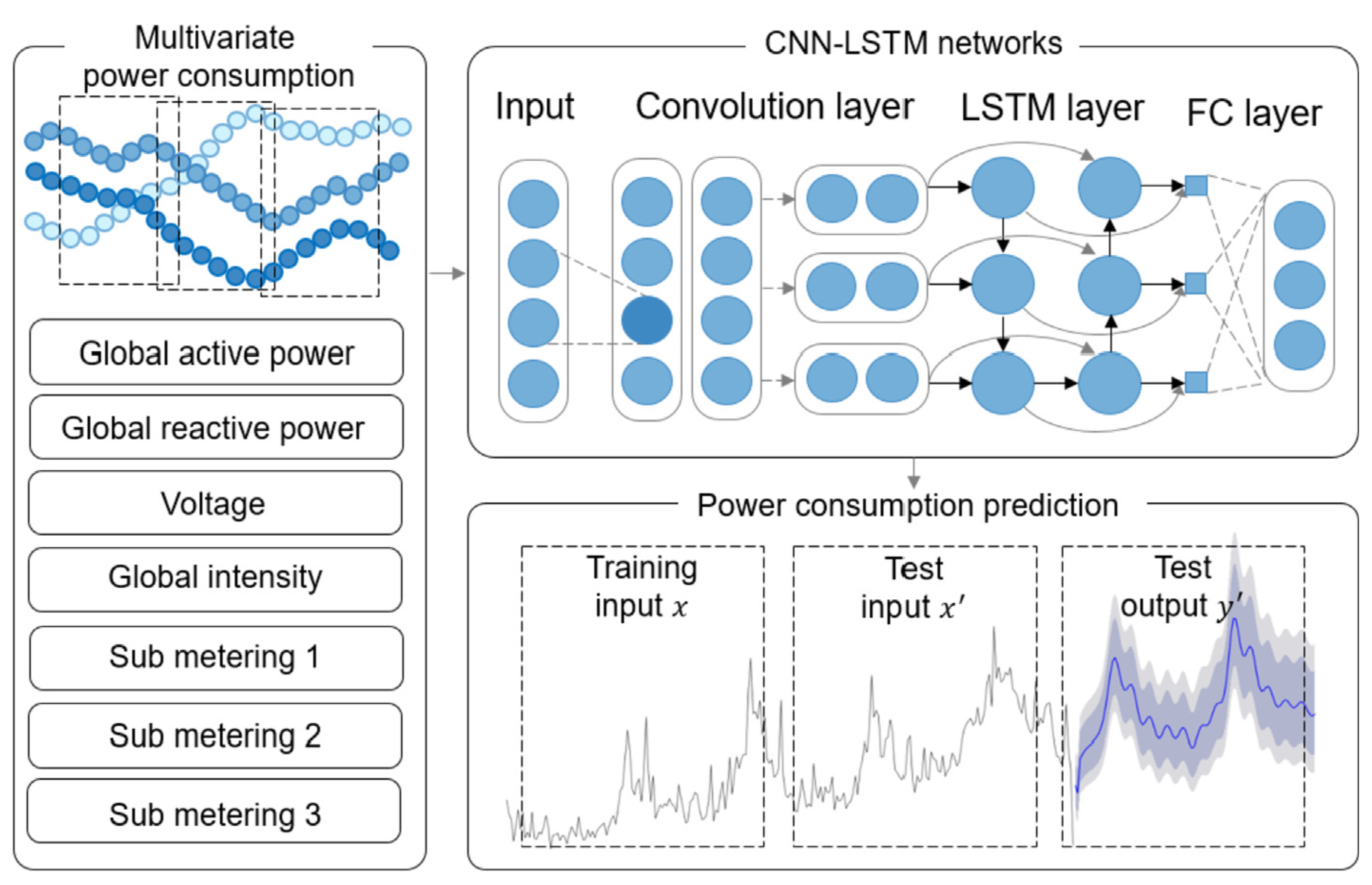}
    \caption{KimTaeYoung model (extracted from \cite{KimTaeYoung}).}
    \label{fig:KimTaeYoung}
\end{figure}

This is a hybrid model combining one-dimensional convolutional layers with LSTM recurrent layers (\Cref{fig:KimTaeYoung}). The structure followed is very similar to the KhanZulfiqar model but changing the GRU layers to one LSTM layer. This model is composed by two one-dimensional convolutions with max pooling layers in between. This first block allows the model to obtain features and improve the quality of them with the pooling operation. The LSTM layer extracts the temporal dependencies from the temporal features obtained previously. The usage of LSTM layers on the one hand prevents gradient explosions, on the other hand it adds computational complexity over the GRU layers.

\subsection{KongZhengmin}
\label{subsection:KongZhengmin}

The KongZhengmin model \cite{KongZhengmin} is designed to predict the remaining useful life from industrial equipment by analysing its degradation.

\begin{figure}[!hbt]
    \centering
    \includegraphics[width=\textwidth]{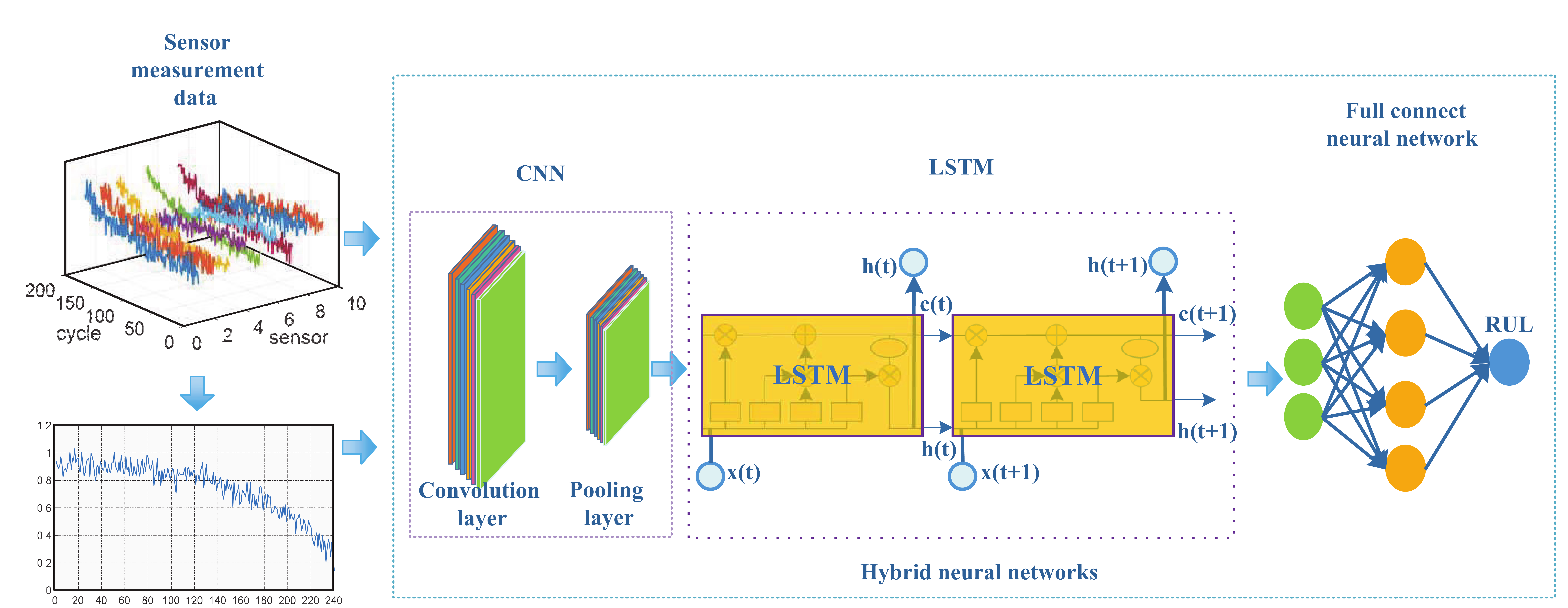}
    \caption{KongZhengmin model (extracted from \cite{KongZhengmin}).}
    \label{fig:KongZhengmin}
\end{figure}

This model (\Cref{fig:KongZhengmin}) is composed by one convolutional layer and two LSTM layers. The one-dimensional convolutional layer followed by the pooling layer extracts temporal features from the time series. In this hybrid model two LSTM layers are applied in order to extract long-term dependencies from the data. This enables the model to gather information from long-lasting pattern.

\subsection{LihOhShu}
\label{subsection:LihOhShu}

The LihOhShu model \cite{LihOhShu} is designed to classify four types of cardiac events from electrocardiograms.

\begin{figure}[!hbt]
    \centering
    \includegraphics[width=\textwidth]{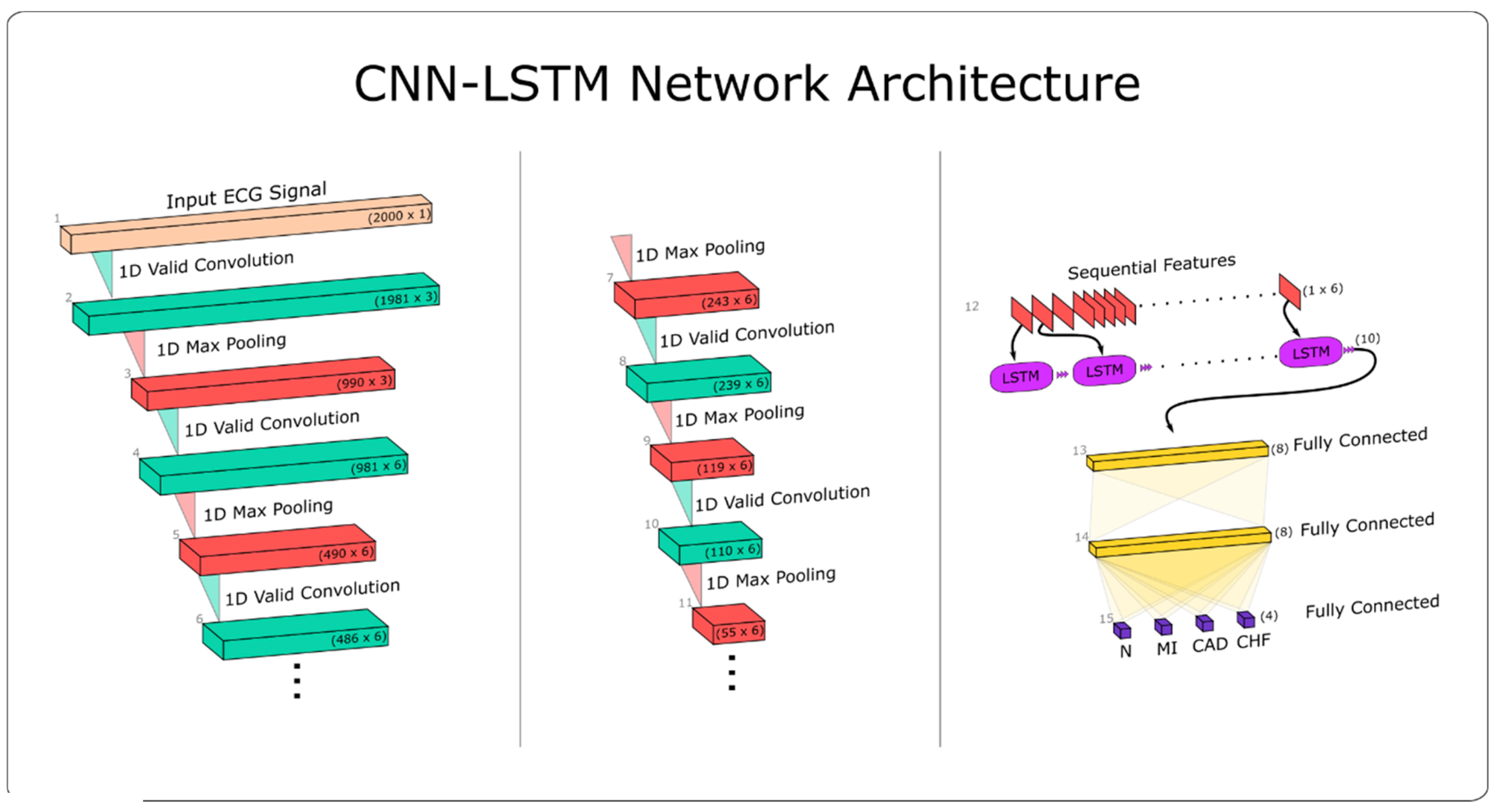}
    \caption{LihOhShu model (extracted from \cite{LihOhShu}).}
    \label{fig:LihOhShu}
\end{figure}

The model presented in \Cref{fig:LihOhShu} consists of 5 convolutional layers with max pooling followed by a LSTM layer. This hybrid model has a very strong convolutional component by using a high number of these layers. The purpose of this network is to classify 4 types of cardiac events. This is a complex task and needs to have high quality features extracted, therefore the model needs to have a higher number of convolutions. After the convolutional block a LSTM layer is used to extract the temporal dependencies.

\subsection{OhShuLih}
\label{subsection:OhShuLih}

The OhShuLih model \cite{OhShuLih} is designed to detect arrhythmia from electrocardiograms.

\begin{figure}[!hbt]
    \centering
    \includegraphics[width=\textwidth]{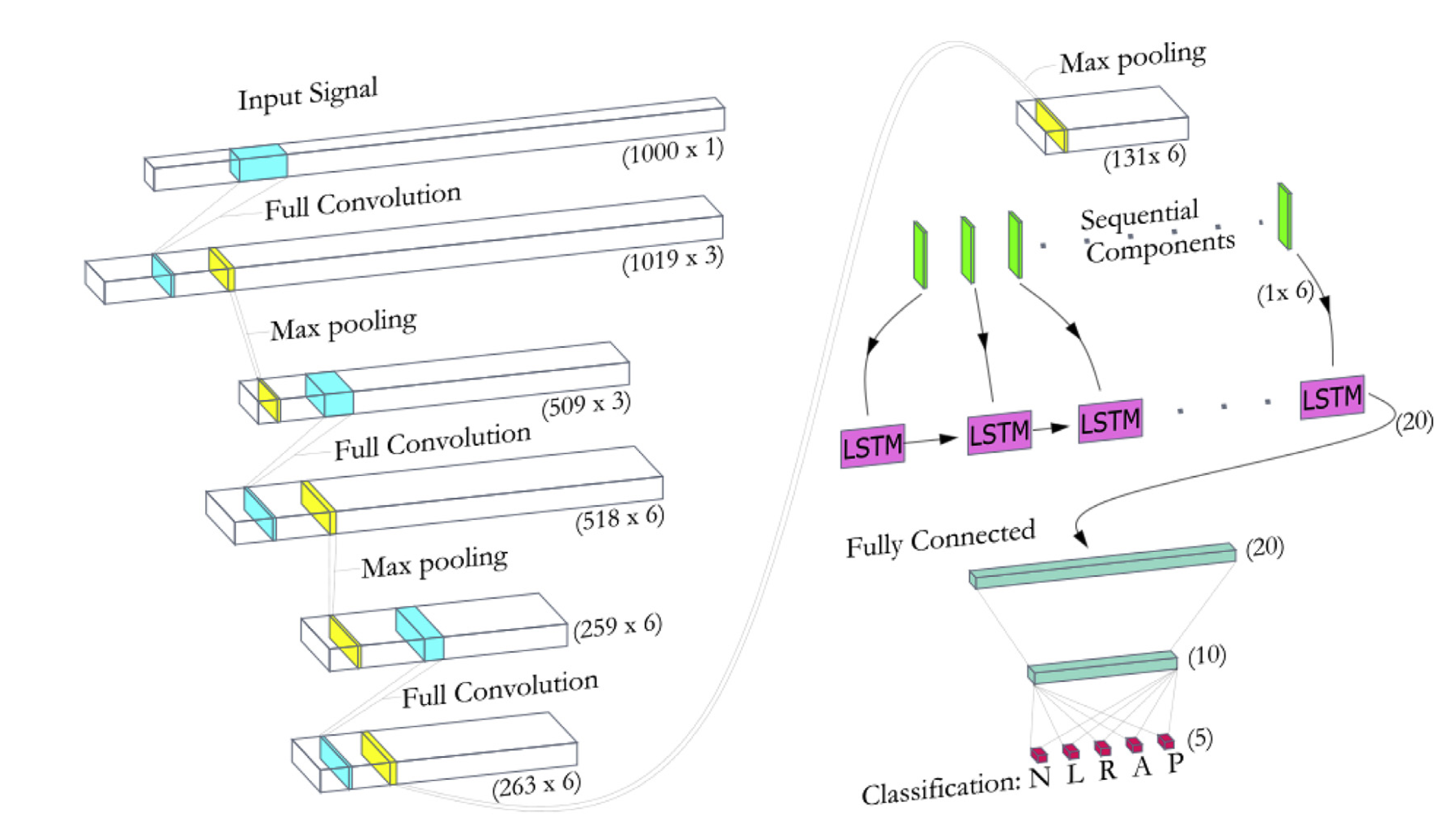}
    \caption{OhShuLih model (extracted from \cite{OhShuLih}).}
    \label{fig:OhShuLih}
\end{figure}

The model described in \Cref{fig:OhShuLih} is composed by 3 full convolutions. The usage of full convolutions instead of valid convolutions is due to the fact that the time series are already zero-padded before entering the model. Between the convolutional layers a max pooling operation is included to increase the quality of the features extracted. Finally a LSTM layer is added to obtain the temporal dependencies from the data.

\subsection{ShiHaotian}
\label{subsection:ShiHaotian}

The ShiHaotian model \cite{ShiHaotian} is designed to detect arrhythmia from electrocardiograms.

\begin{figure}[!hbt]
    \centering
    \includegraphics[scale=0.04]{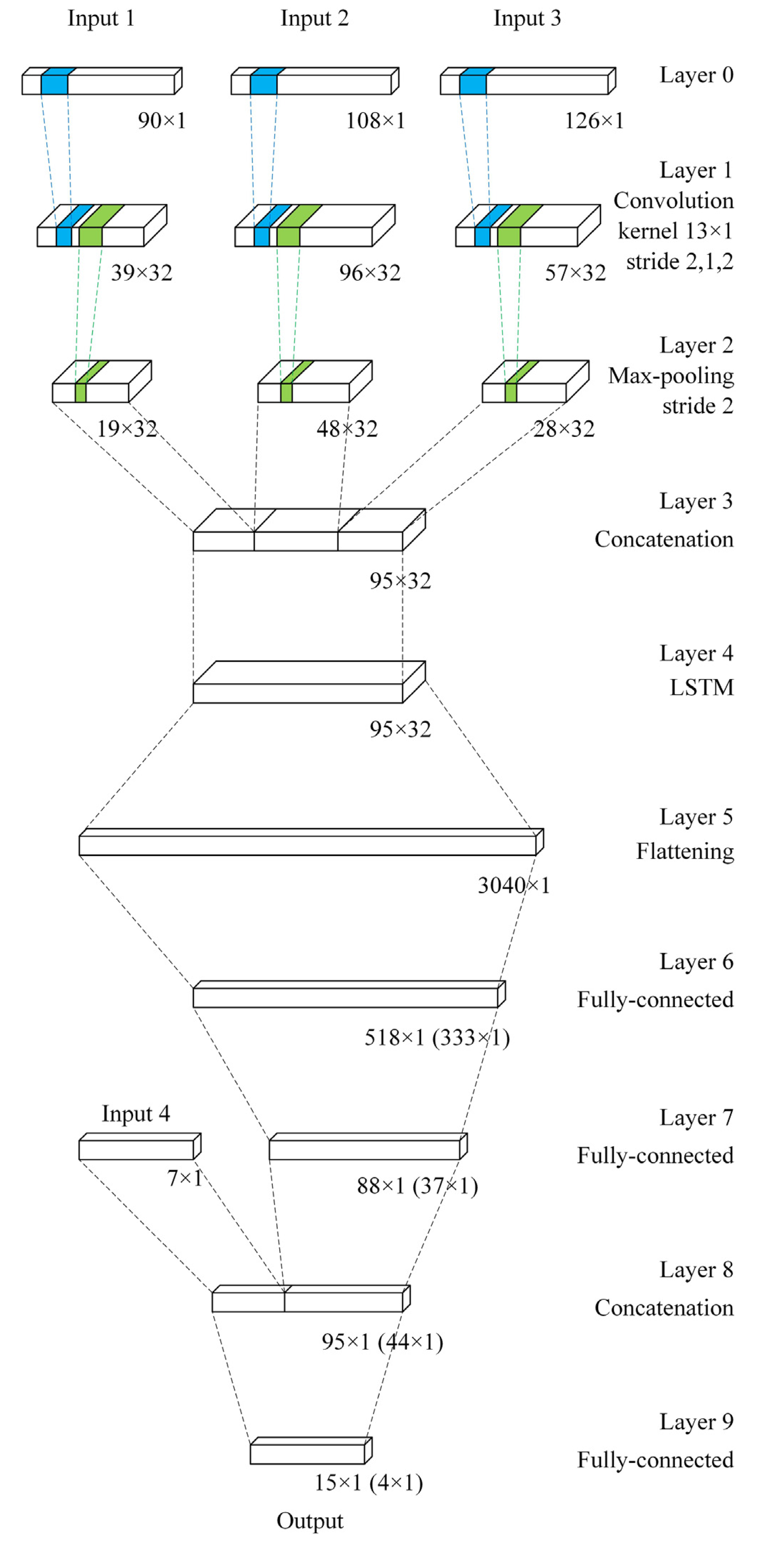}
    \caption{ShiHaotian model (extracted from \cite{ShiHaotian}).}
    \label{fig:ShiHaotian}
\end{figure}

The model presented in \Cref{fig:ShiHaotian} is a hybrid CNN-LSTM model. The network has three inputs, having one independent convolution and max pooling for each of them. The three inputs are motivated by the dataset, in order to have the classification three different section of the electrocardiogram are obtained and passed to the model. After the convolution and pooling the result is concatenated in a single feature array and passed to an LSTM layer. In the last section of the fully connected layers a fourth input is concatenated, containing this input seven manually computed features. This model is a suitable model to use for extracting features in a parallel fashion to obtain different representations of the data at the same time. The addition of another input at the end of the model makes it suitable for incorporating expert knowledge inside the neural network.

\subsection{WangKejun}
\label{subsection:WangKejun}

The WangKejun model \cite{WangKejun} is designed to predict the energy generated from a photovoltaic power plant based on the weather data and solar radiation values. 

\begin{figure}[!hbt]
    \centering
    \includegraphics[width=\textwidth]{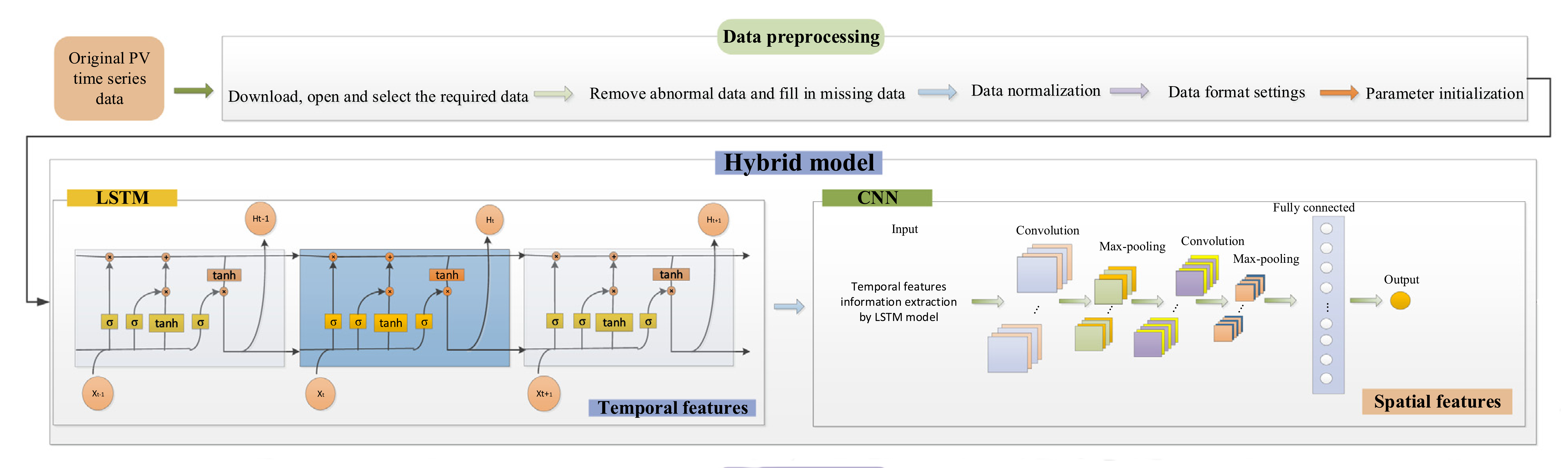}
    \caption{WangKejun model (extracted from \cite{WangKejun}).}
    \label{fig:WangKejun}
\end{figure}

The model presented in \Cref{fig:WangKejun} is a hybrid model combining one-dimensional convolutions with LSTM layers. The way of extracting the features in this model is by first extracting the temporal features with two LSTM layers and then applying two convolutional layers for spatial feature extraction. The choice of applying first the LSTM layers is based on the empirical results obtained. In the paper the authors compare a CNN-LSTM architecture against a LSTM-CNN architecture, obtaining as final results that the LSTM-CNN model obtains better results in their problem. Therefore the temporal features in their dataset are more important than the spatial features, being this model a suitable option for this kind o data.

\subsection{WeiXiaoyan}
\label{subsection:WeiXiaoyan}

The WeiXiaoyan model \cite{WeiXiaoyan} is designed to predict epileptic seizures using electroencephalograms. 

\begin{figure}[!hbt]
    \centering
    \includegraphics[scale=0.05]{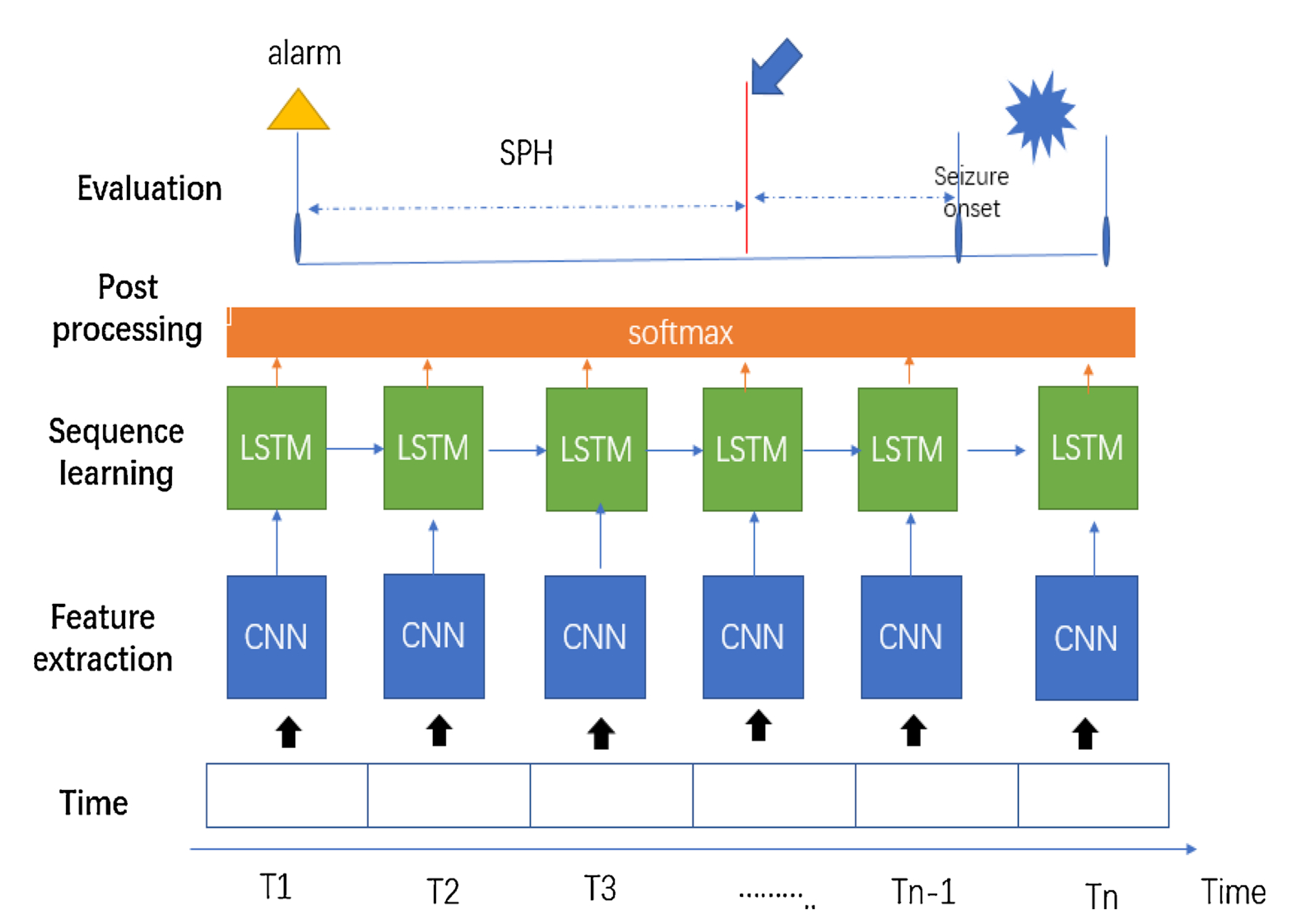}
    \caption{WeiXiaoyan model (extracted from \cite{WeiXiaoyan}).}
    \label{fig:WeiXiaoyan}
\end{figure}

The model presented in \Cref{fig:WeiXiaoyan} is a hybrid CNN-LSTM model. This architecture is composed by 5 convolutional blocks which are composed by a one-dimensional convolution, followed by a leaky ReLU activation function, a max pooling operation and a batch normalisation layer. After these convolutional blocks we find an LSTM layer, followed by a batch normalisation layer and another LSTM layer. This model ensures that after each convolution the distribution of the output data continues having zero as mean and unit variance, reducing the changes of the underlying distribution. The LSTM layers extract the temporal features, having again a batch normalisation to ensure that the distribution is not changing. The batch normalisation is used in this architecture to remove the necessity of the dropout layers to prevent overfitting of the weights. This also allows the usage of a higher learning rate in order to train the model faster converging in less epochs.

\subsection{YaoQihang}
\label{subsection:YaoQihang}

The YaoQihang model \cite{YaoQihang} is designed to detect arrhythmia using electrocardiograms.

\begin{figure}[!hbt]
    \centering
    \includegraphics[scale=0.05]{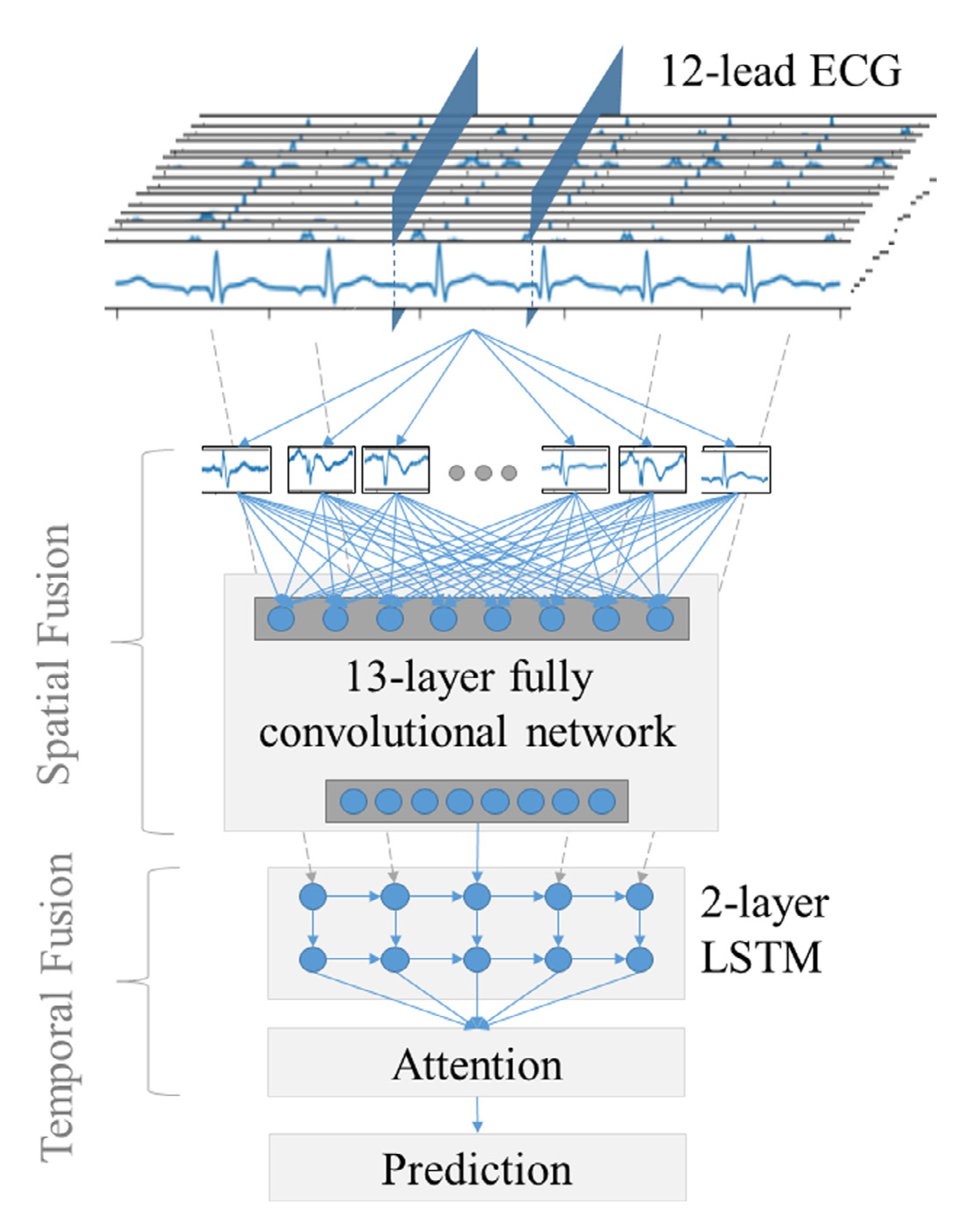}
    \caption{YaoQihang model (extracted from \cite{YaoQihang}).}
    \label{fig:YaoQihang}
\end{figure}

The model described in \Cref{fig:YaoQihang} is a hybrid model combining CNN and LSTM layers. This architecture is composed by 13 convolutional layers, adding a batch normalization layer, a ReLU activation and a max pooling operation on each convolution block. This architecture builds a solid spatial feature extraction module. The convolutions are followed by two LSTM layers which extract the temporal features. Finally an attention block is used, composed by two fully connected layers along with a hyperbolic tangent function. The attention block is not used in the network implementation as it is a problem-dependant choice.

\subsection{YiboGao}
\label{subsection:YiboGao}

The YiboGao model \cite{YiboGao} is designed to detect atrial fibrillation using electrocardiograms.

\begin{figure}[!hbt]
    \centering
    \includegraphics[width=\textwidth]{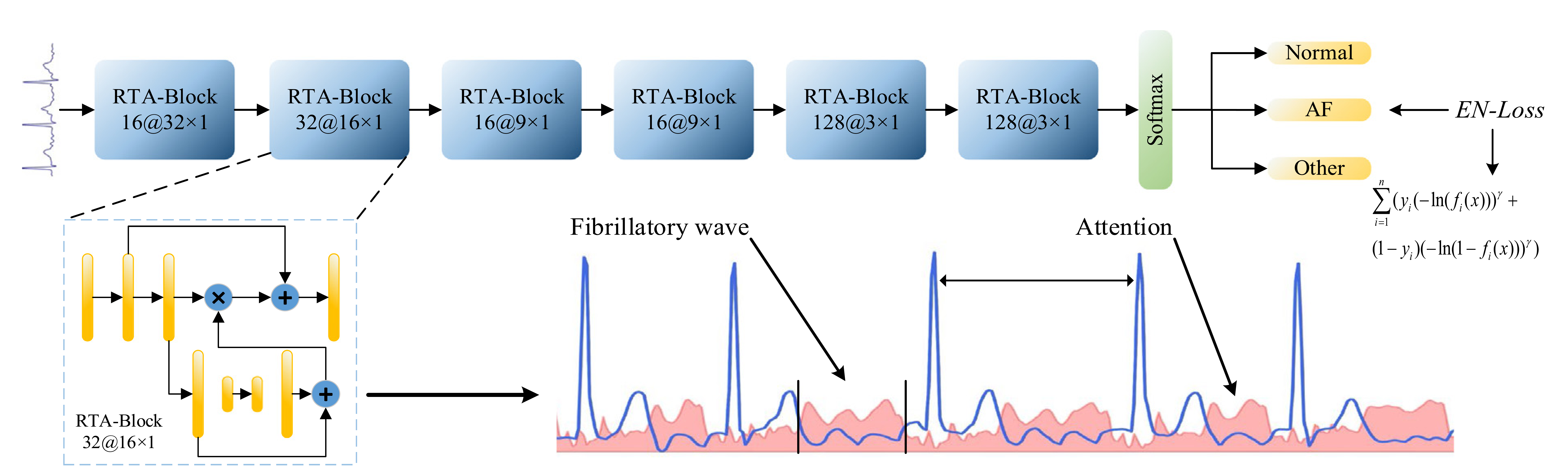}
    \caption{YiboGao model (extracted from \cite{YiboGao}).}
    \label{fig:YiboGao}
\end{figure}

The model presented in \Cref{fig:YiboGao} is a convolutional model composed by Residual-based Temporal Attention blocks, abbreviated as RTA blocks.

\begin{figure}[!hbt]
    \centering
    \includegraphics[width=\textwidth]{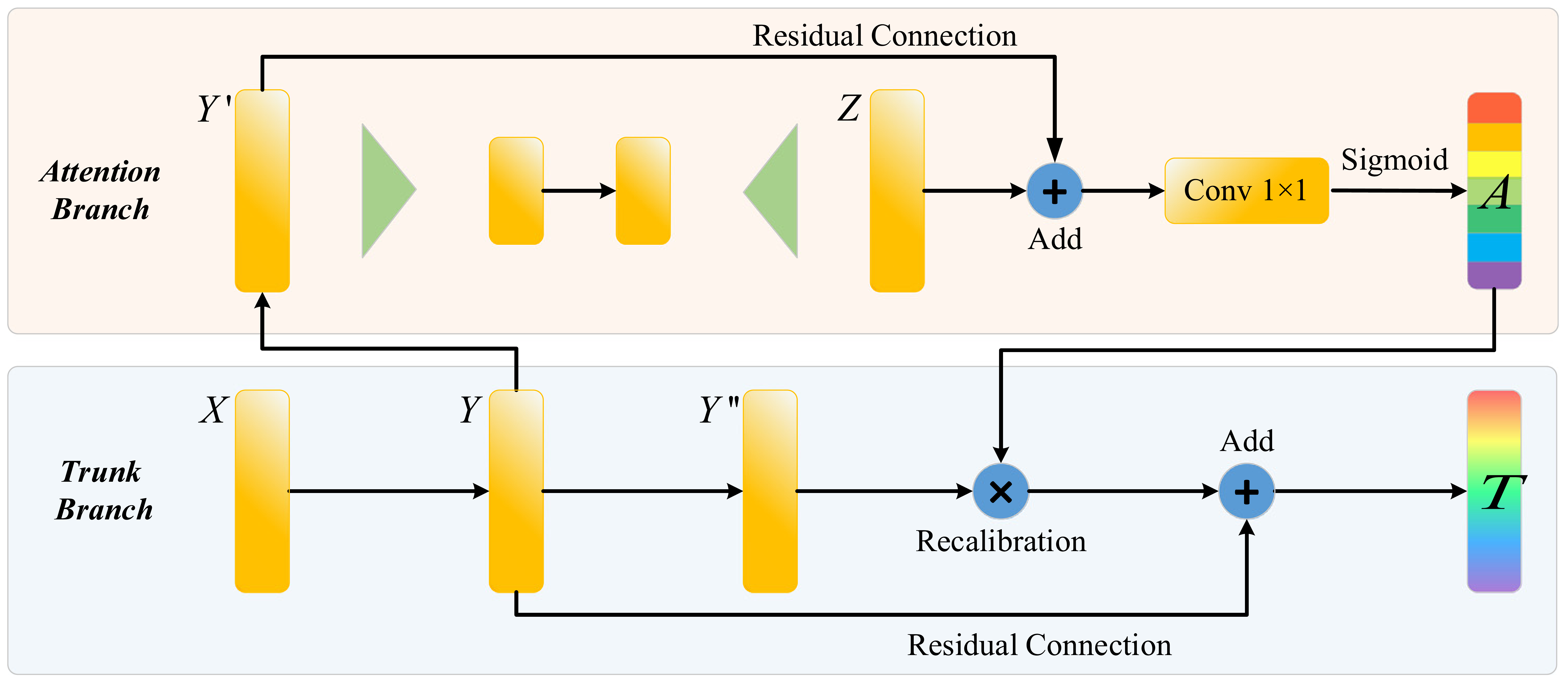}
    \caption{YiboGao RTA block (extracted from \cite{YiboGao}).}
    \label{fig:YiboGao2}
\end{figure}

In the author's proposal they emphasise that their data could have noise in the signal, so the feature extraction might be more difficult. For filtering this unwanted behaviour they define the RTA block seen in \Cref{fig:YiboGao2}. The RTA block is composed by two different branches: the trunk branch and the attention branch. The attention branch serves as a feature selector to generate temporal attention weights whereas the trunk branch processes these features to enhance the relevant features and supress the non-informative ones.

The RTA block is composed by a set of one-dimensional convolutions, batch normalization layers and max pooling operations. The block includes also upsampling and cropping procedures to focus the attention on the relevant features.

\subsection{YildirimOzal}
\label{subsection:YildirimOzal}

The YildirimOzal model \cite{YildirimOzal} is designed to detect arrhythmia using electrocardiograms.

\begin{figure}[!hbt]
    \centering
    \includegraphics[scale=0.05]{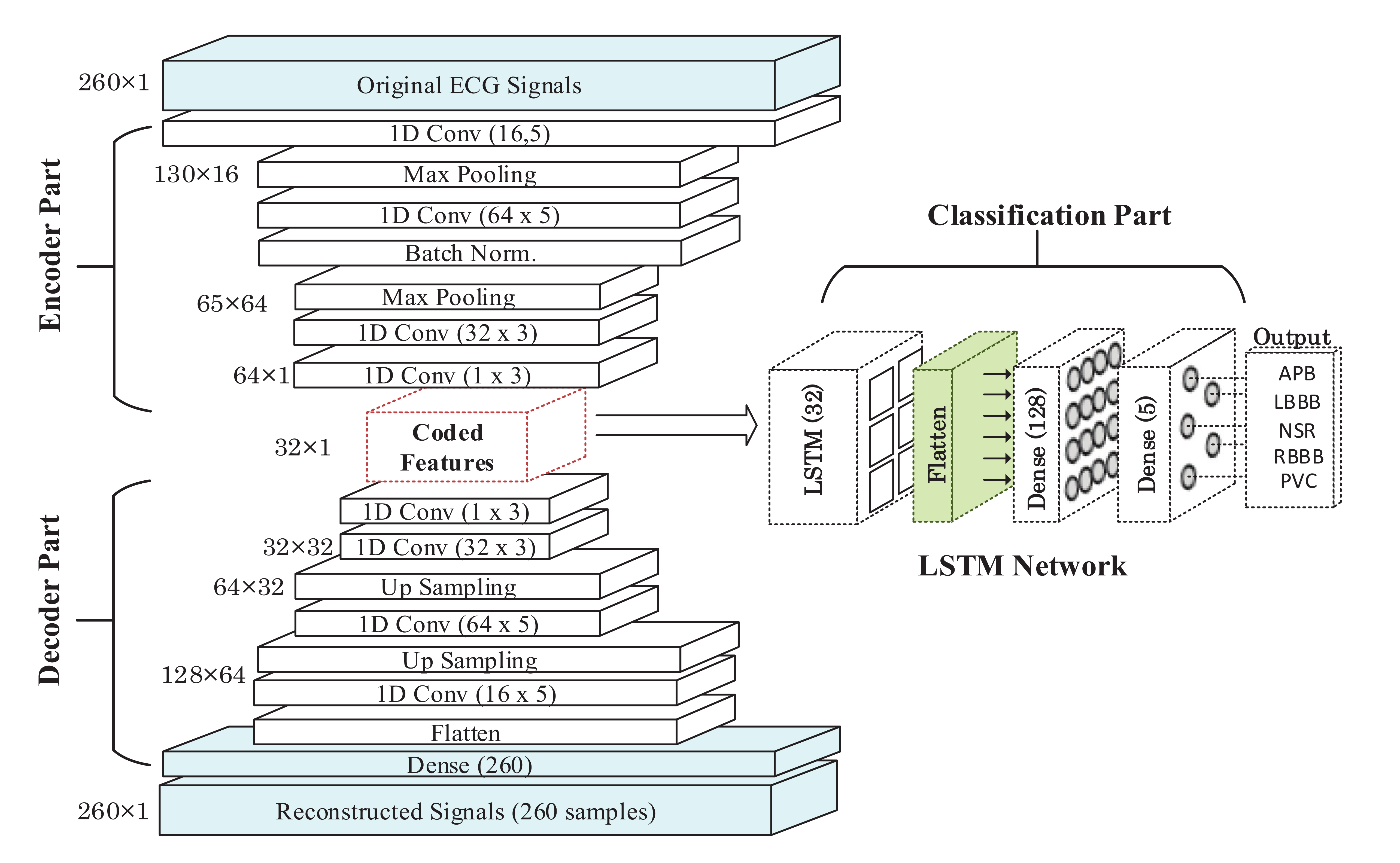}
    \caption{YildirimOzal model (extracted from \cite{YildirimOzal}).}
    \label{fig:YildirimOzal}
\end{figure}

This model (\Cref{fig:YildirimOzal}) is an convolutional autoencoder combined with a classification module with an LSTM layer. This autoencoder learns a coded representation of the input signal by obtaining features from it and reconstructing the original signal from the encoded features. After training the encoder the classification module is used only with the encoding architecture. The coded features are passed to the classification module which is composed by an LSTM layer to extract temporal dependencies from the spatial features obtained by the convolutional autoencoder.

\subsection{ZhangJin}
\label{subsection:ZhangJin}

The ZhangJin model \cite{ZhangJin} is designed to detect arrhythmia using electrocardiograms. 

\begin{figure}[!hbt]
    \centering
    \includegraphics[scale=0.07]{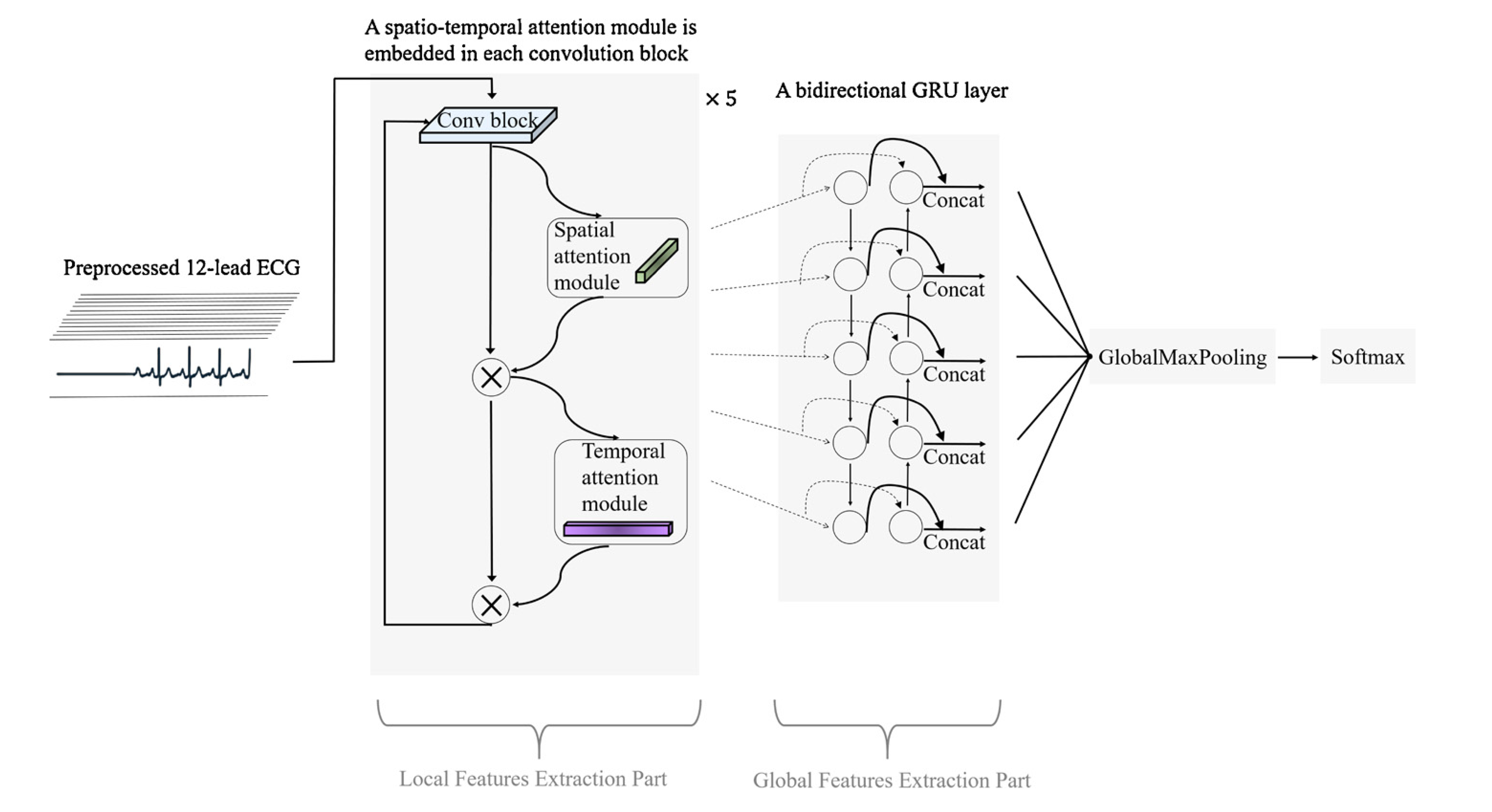}
    \caption{ZhangJin model (extracted from \cite{ZhangJin}).}
    \label{fig:ZhangJin}
\end{figure}

The model presented in \Cref{fig:ZhangJin} is a hybrid CNN-GRU model. The architecture starts with a convolutional layer followed by spatial and temporal attention modules. 

\begin{figure}[!hbt]
    \centering
    \includegraphics[scale=0.07]{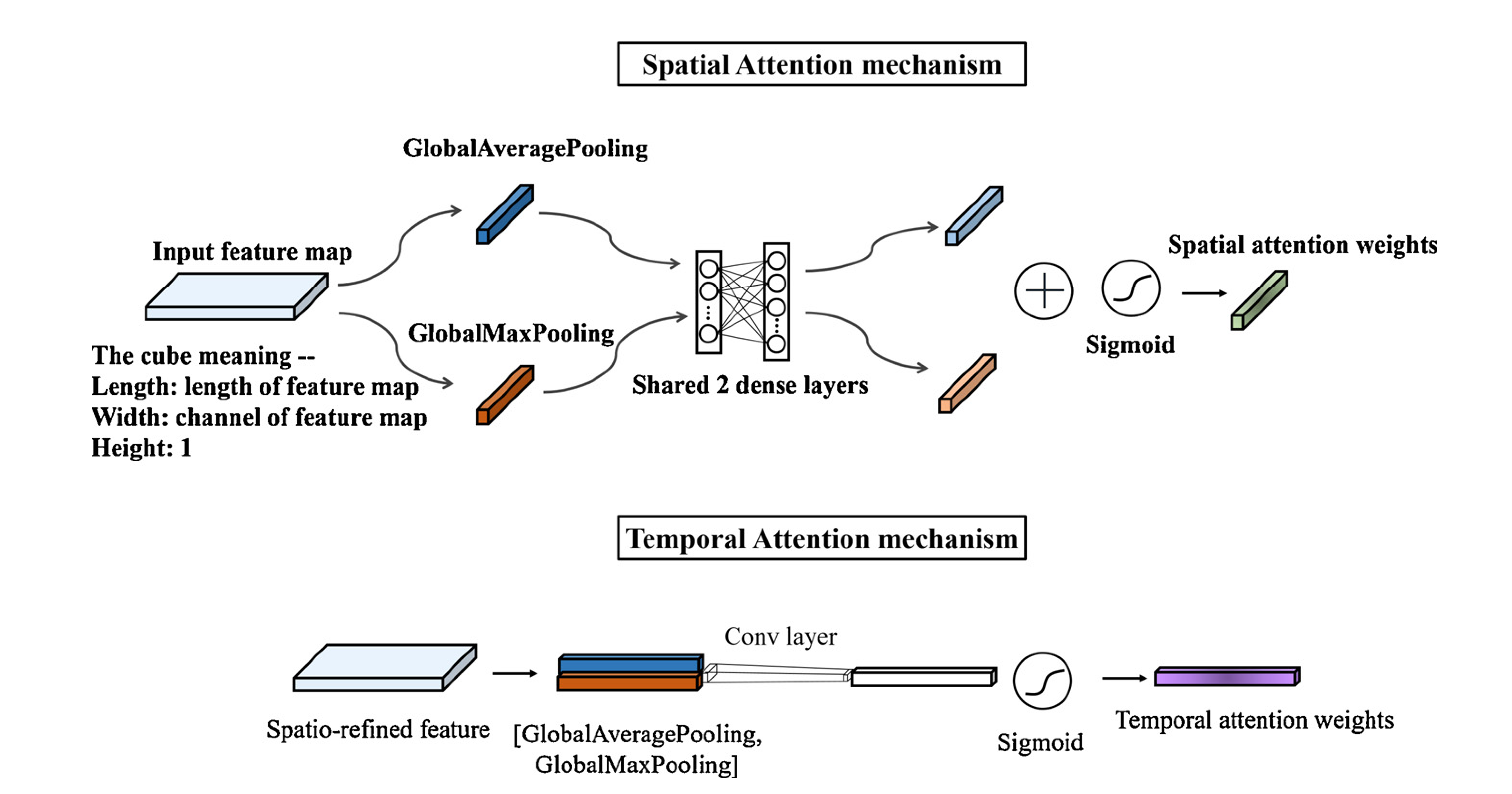}
    \caption{ZhangJin spatial and temporal attention modules (extracted from \cite{ZhangJin}).}
    \label{fig:ZhangJin2}
\end{figure}

The modules from \Cref{fig:ZhangJin2} serve as a tool for this model to extract representative local spatio-temporal features. The spatial attention mechanism weights the feature map to focus on the most important features. After this, the temporal attention module weights the temporal segments from the important features to obtain the most relevant temporal intervals from the relevant features. All this procedure uses only convolutional and fully connected layers.

\subsection{ZhengZhenyu}
\label{subsection:ZhengZhenyu}

The ZhengZhenyu model \cite{ZhengZhenyu} is designed to detect arrhythmia using electrocardiograms.

\begin{figure}[!hbt]
    \centering
    \includegraphics[scale=0.04]{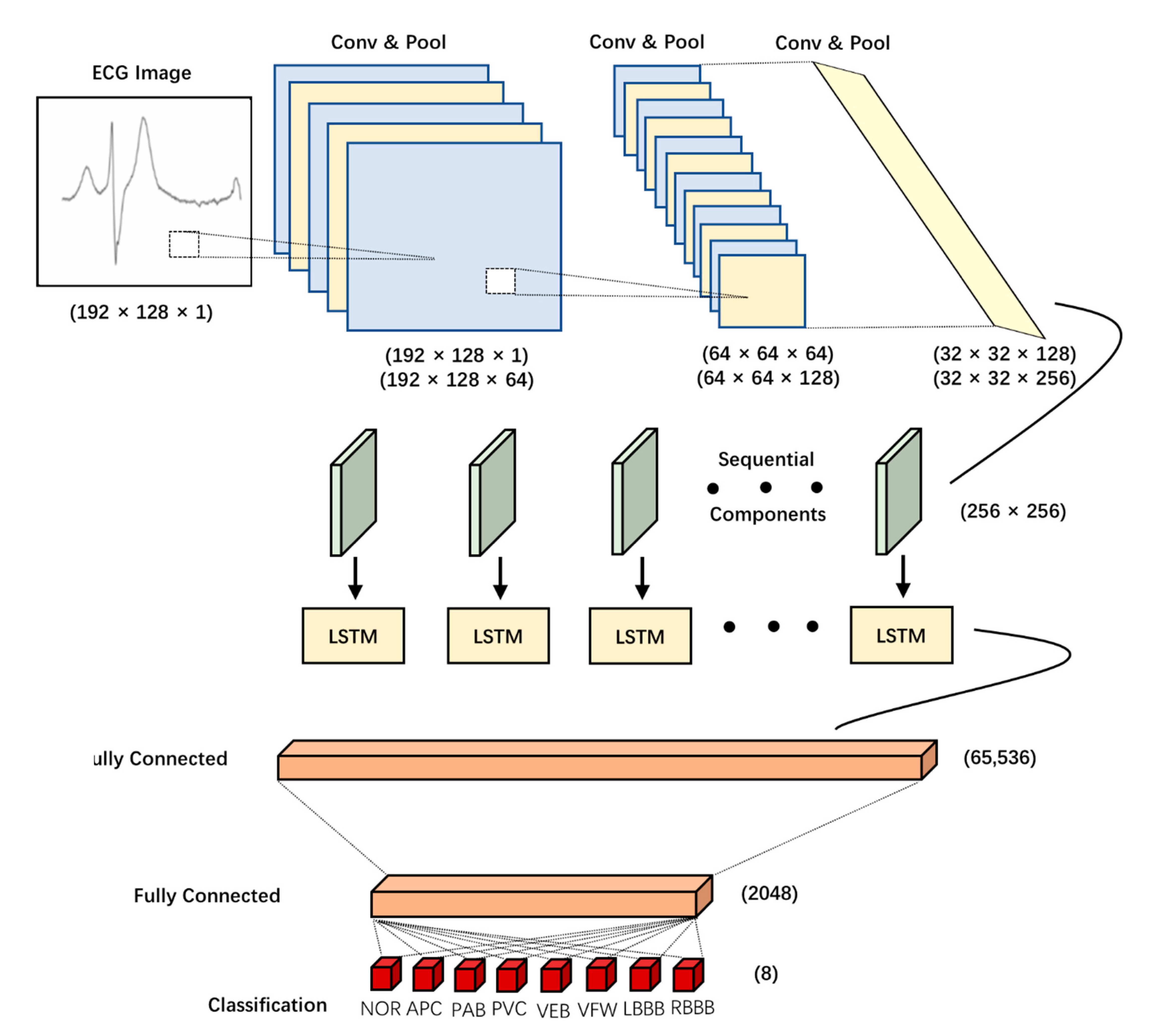}
    \caption{ZhengZhenyu model (extracted from \cite{ZhengZhenyu}).}
    \label{fig:ZhengZhenyu}
\end{figure}

The model presented in \Cref{fig:ZhengZhenyu} is a hybrid CNN-LSTM model. The presented architecture contains 3 convolutional blocks composed by 2 one-dimensional convolutions followed by two batch normalization operations and a max pooling operation. Finally the output of these three convolutional blocks is passed to a LSTM layer to extract the temporal features.

\subsection{HongTan}
\label{subsection:HongTan}

The HongTan model \cite{HongTan} is designed to detect arrhythmia using electrocardiograms.

\begin{figure}[!hbt]
    \centering
    \includegraphics[scale=0.045]{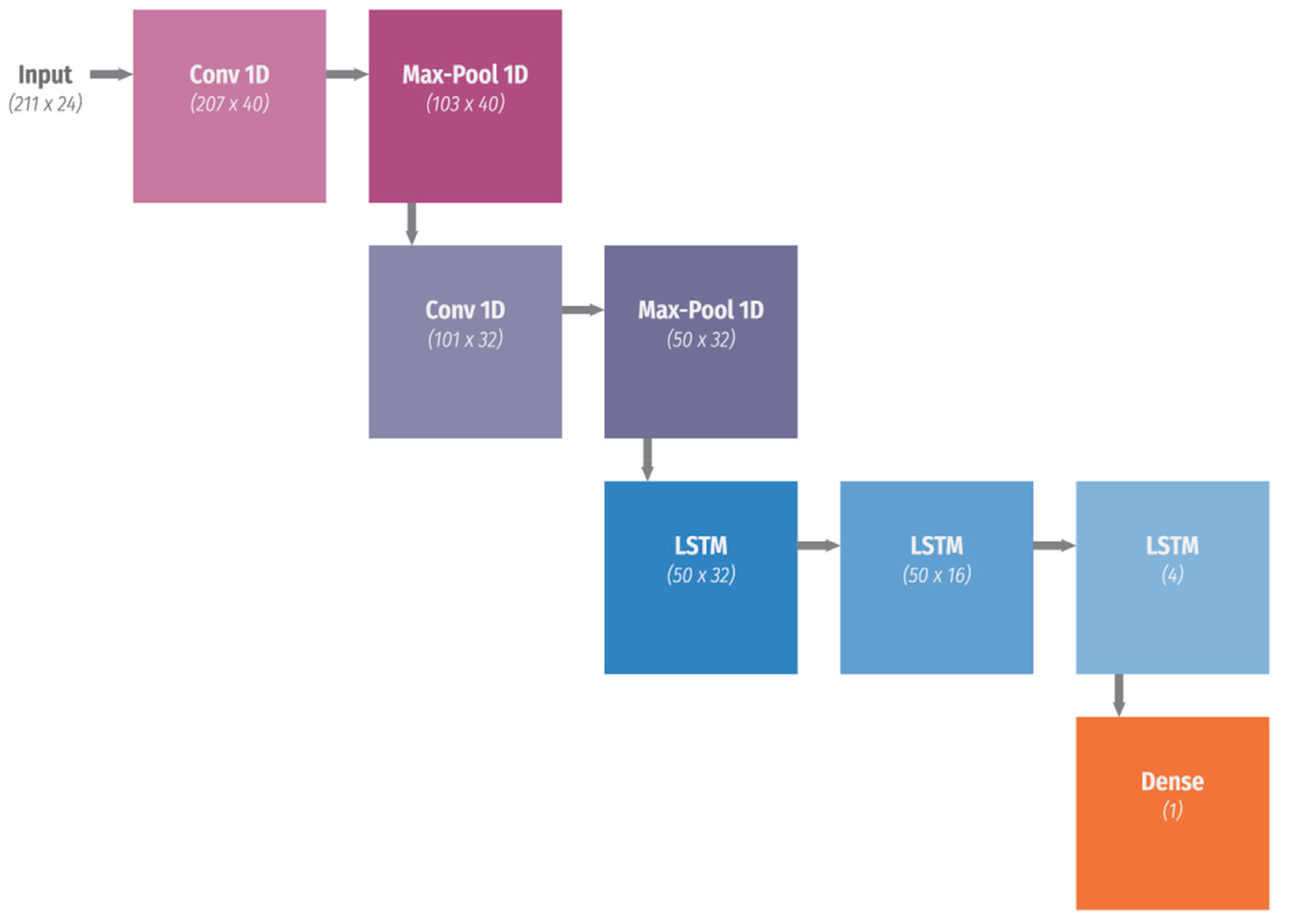}
    \caption{HongTan model (extracted from \cite{HongTan}).}
    \label{fig:HongTan}
\end{figure}

The model presented in \Cref{fig:HongTan} is a hybrid CNN-LSTM model. This model has two one-dimensional convolutions followed by max pooling operations. The main characteristic of this model is the usage of three LSTM layers for temporal feature extraction. Using three LSTM layers forces this architecture to obtain characteristics from long patterns.

\section{Experimental studies}
\label{appendix:experimental-studies}

A time series is a sequence of values, indexed by timestamps, taken sequentially in time \cite{Hamilton2020}. They can be represented by a list of values $\mathcal{X} = (x_{T_0}, x_{T_2}, ..., x_{T_n})$ indexed by the temporal index $\mathcal{T} = \{T_1, T_2, ..., T_n\}$ where $x_{T_i}\in \mathbb{R}$. The general definition of a single-valued time series can be extended to a multi-valued time series arranging a set of single-valued time series by columns. Therefore a $K$-valued time series can be noted as $\mathcal{X}^K = (\mathcal{X}_{1}, \mathcal{X}_{2}, ..., \mathcal{X}_{K})$. 
A $k$-valued time series is a set of single-valued time series arranged by columns.

\begin{equation}
    \scriptsize
    \mathcal{X} = (\mathcal{X}_1, \mathcal{X}_2, ..., \mathcal{X}_k) = 
    \begin{pmatrix}
        x_{T_1}^1 & x_{T_1}^2 & \cdots & x_{T_1}^k \\
        x_{T_2}^1 & x_{T_2}^2 & \cdots & x_{T_2}^k \\
        \vdots & \cdots & \ddots & \vdots \\
        x_{T_n}^1 & x_{T_n}^2 & \cdots & x_{T_n}^k \\
    \end{pmatrix}
    \ where \  \mathcal{X}\in \mathbb{M}_{n\times k}(\mathbb{R}).
\end{equation}

When dealing with time series several problems can be faced such as classification, forecasting or anomaly detection. Time series classification consists in determining the class of each time step of the time series or the class of a batch of samples from a time series. Time series forecasting is the problem of predicting the behaviour of future time steps. Finally, time series anomaly detection consists in scoring each sample or batch of samples of a time series, being the higher the score the more anomalous the sample. In all cases, the relevance of the extracted spatio-temporal features is highly significant, because if irrelevant features are extracted it is not possible to generalise properly and overfitting occurs.

In this paper, the three types of time series prediction problems are carried out: forecasting, classification and anomaly detection. The goals of this study are to create a new CNN-RNN model leveraging the characteristics of the proposed library and study experimentally the performance of the implemented architectures. In this section, each problem proposed in this work is detailed in order to provide additional information about the data and their preprocessing.

\subsection{Forecasting: Spanish Digital Seismic Network} 
\label{appendix:IGN}

The Spanish Digital Seismic Network is a seismograph network distributed across the territory of Spain in order to watch for seismic movements and alert the population in case of earthquakes, tsunami or volcanic eruptions. Each seismograph records data on three different components (HHE, HHN and HHZ) at a sampling rate of 100Hz. The Spanish Digital Seismic Network provides this data on a daily basis. 

The purpose of this experimental study is to predict the future $n$ timesteps of a given component in a given seismograph. Specifically, the dataset employed in this study corresponds to the data of the HHE component of a seismograph located at Adamuz (C\'ordoba, Spain) using the records of the 2nd of January of 2021. This dataset is therefore composed of 1 dimension and 8640000 timesteps.

\begin{figure}[!hbtp]
    \centering
    \includegraphics[width=0.5\linewidth]{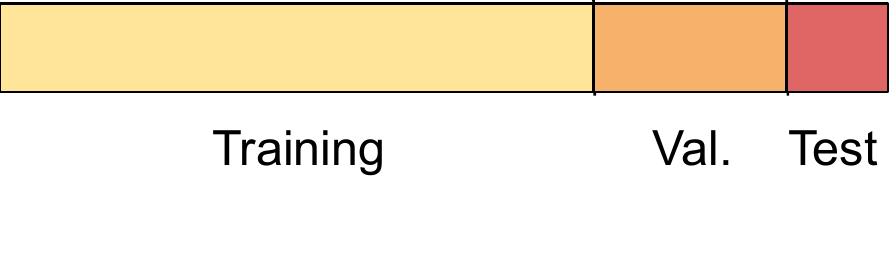}
    \caption{Time series split into training, validation and test sets for this experimental study. The sizes of each set is 70\%, 20\% and 10\% of the total size respectively.}
    \label{fig:IGN_split}
\end{figure}

Data have been slightly preprocessed for this study. Specifically, a smoothing process using the moving average has been employed. The size of the windows to compute the moving average is $w=50$. This process has been iteratively applied five times for further smoothing. After that, the time series has been split into the training, validation and test sets. In this case, the first 70\% of timesteps have been used for training, the following 20\% have been used for the validation set and the remaining 10\% of data are employed for test. A graphical representation of this partitioning is shown in \Cref{fig:IGN_split}.

Finally, to properly train the deep learning models included in the library it is required to provide the data as batches of instances $x_1, x_2, \dots, x_n$ of a fixed size. Let us suppose the size of the input data is $i$ and the size of the output data (the forecasting) is $o$. Each instance $x_k$ is composed by the timesteps from $T_k$ to $T_{k+i}$ as the input data and the timesteps from $T_{k+i+1}$ to $T_{k+i+o}$ as the labels to perform the training. In this example $i = 1000$ and $o = 50$, so this leads us to a final dataset of approximately 6 million instances with 1000 variables each.

The training process has been followed using the Adam optimiser with a batch size of 256. The mean squared error has been used as the loss function. The training process was configured to run for 150 epochs together with an early stopping procedure with patience 2 and delta 0.

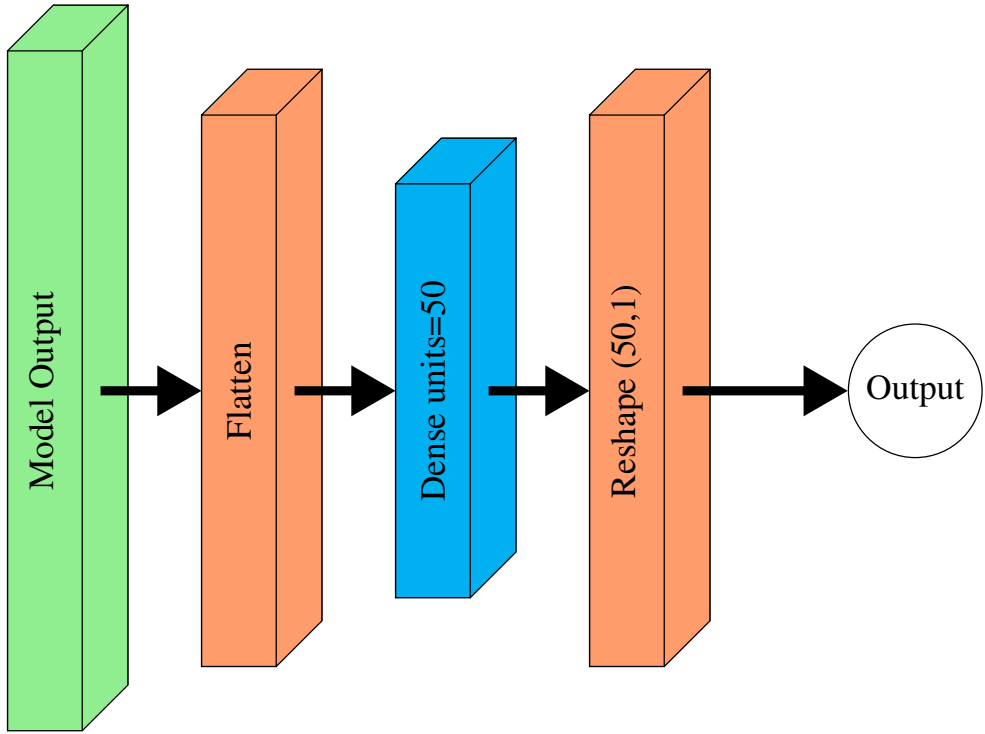
\begin{figure}[!hbt]
    \centering
    \resizebox{.8\linewidth}{!}{
        \begin{tikzpicture}
          \node[input](input){\rotatebox{90}{Model Output}};
          
          \node[conv,minimum height=6cm,right=1.3cm of input](flatten){\rotatebox{90}{Flatten}};
          
          \node[dense,minimum height=4.5cm,right=1.3cm of flatten](dense1){\rotatebox{90}{Dense units=50}};
          
          \node[conv,minimum height=6cm,right=1.3cm of dense1](reshape){\rotatebox{90}{Reshape (50,1)}};
          
          \node[circle,draw,radius=1cm,right=2cm of reshape](out){Output};
        
          \draw [-triangle 60,link] ([xshift=0.2cm]input.east) -- (flatten.west);
          \draw [-triangle 60,link] ([xshift=0.2cm]flatten.east) -- (dense1.west);
          \draw [-triangle 60,link] ([xshift=0.2cm]dense1.east) -- (reshape.west);
          \draw [-triangle 60,link] ([xshift=0.2cm]reshape.east) -- (out.west);
        
        \end{tikzpicture}
     }
    \caption{Specialisation module for the time series forecasting task.}
    \label{fig:IGN_spec}
\end{figure}

The specialisation module employed in this problem is described in \Cref{fig:IGN_spec}. The first operation is  to flatten the output of the model. After that a dense layer of 50 ReLU units is employed to perform the forecasting. Note that the size of 50 is equal to the output size of the problem. Finally, the data is reshaped in a matrix fashion to compare it to the real values.

\subsection{Classification: Arrhythmia detection} 
\label{appendix:arrythmia}

The well-known MIT-BIH dataset \cite{MIT_BIH} was begun to be distributed in 1980 and it is the first generally available set of standard test material for the evaluation of arrhythmia detectors. The dataset contains 48 half-hour excerpts of two-channel ambulatory ECG recordings. The sampling rate was 360Hz and they are accompanied by a set of beat labels rendered at the R peak.

The objective of this study is to classify different segments of this dataset to differentiate between normal beats (N), left bundle branch block (L), right bundle branch block (R), atrial premature beats (A) and premature ventricular contraction (V). The remaining labels of this dataset were discarded.

The ECG segments for this study were generated following the same procedure described in \cite{OhShuLih}. All of them have a fixed size of 1000 timesteps. They were created by assigning the uninterrupted sequences of R peaks of the same type until a new one appeared. Each one begins 99 timesteps before the first R peak and it ends 160 timesteps after the last identified R peak. As the sizes of the segments were very different, those larger than 1000 timesteps were truncated, whereas the shorter ones were padded with zeroes until reaching the target size. Finally, a z-score normalisation has been carried out to eliminate the offset effect and to standardise the ECG signal amplitude. This procedure leads us to a dataset of 16498 segments.

The training process has been followed using the Adam optimiser with a batch size of 256. The categorical cross-entropy has been used as the loss function. The training process was configured to run for 150 epochs together with an early stopping procedure with patience 3 and delta 0.

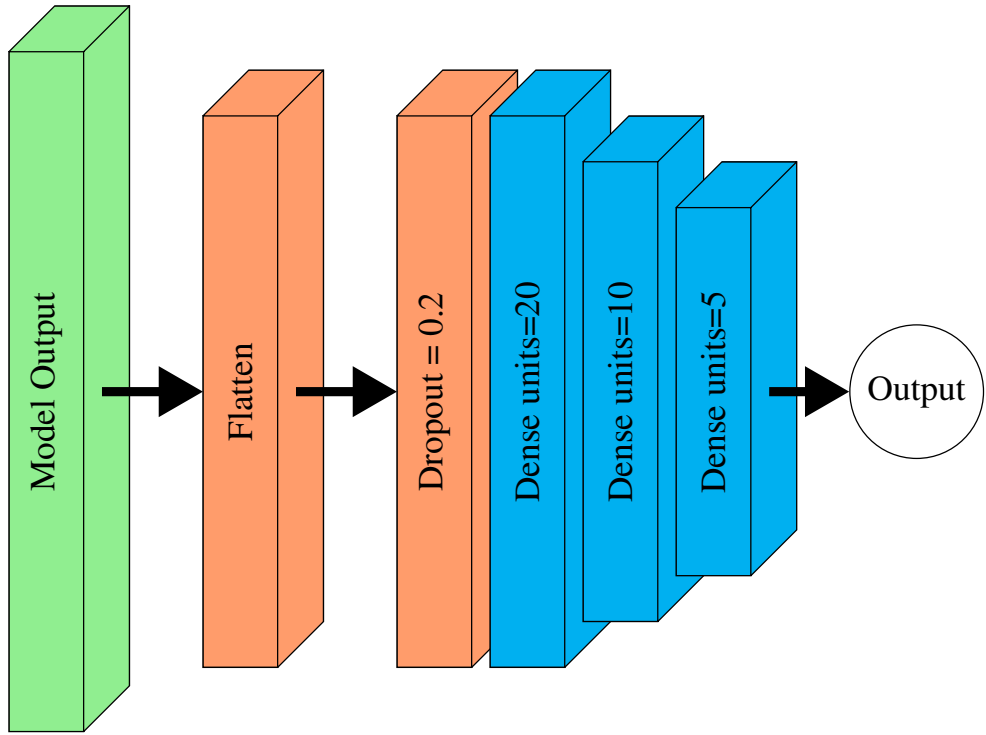
\begin{figure}[!hbt]
    \centering
    \resizebox{.8\linewidth}{!}{
        \begin{tikzpicture}
          \node[input](input){\rotatebox{90}{Model Output}};
          
          \node[conv,minimum height=6cm,right=1.3cm of input](flatten){\rotatebox{90}{Flatten}};
          
          \node[conv,minimum height=6cm,right=1.3cm of flatten](dense1){\rotatebox{90}{Dropout = 0.2}};
          \node[dense,minimum height=6cm,right=0.2cm of dense1](dense2){\rotatebox{90}{Dense units=20}};
          \node[dense,minimum height=5cm,right=0.2cm of dense2](dense3){\rotatebox{90}{Dense units=10}};
          \node[dense,minimum height=4cm,right=0.2cm of dense3](dense4){\rotatebox{90}{Dense units=5}};

          \node[circle,draw,radius=1cm,right=2cm of reshape](out){Output};
        
          \draw [-triangle 60,link] ([xshift=0.2cm]input.east) -- (flatten.west);
          \draw [-triangle 60,link] ([xshift=0.2cm]flatten.east) -- (dense1.west);
          \draw [-triangle 60,link] ([xshift=0.2cm]dense4.east) -- (out.west);
        
        \end{tikzpicture}
     }
    \caption{Specialisation module for the classification task.}
    \label{fig:arrythmia}
\end{figure}

The specialisation module employed in this problem is described in \Cref{fig:arrythmia}. The first operation is to flatten the output of the model. After that a dropout operation is applied in order to better generalise the training procedure. Then several dense layers with ReLU units are applied to train the model, where the last dense layer of five units corresponds to the final output in a softmax format.

\subsection{Anomaly detection: Malicious attacks in network traffic}
\label{appendix:anomaly}

The KDD cup 99 dataset \cite{kddcup99} is the dataset which was proposed in the third International Knowledge Discovery and Data Mining Tools Competition. This dataset defines an intrusion detection problem inside a military network environment, simulating these events.

The dataset has 126 features with 24 different classes corresponding to 23 different types of attacks and one normal class which labels the non-anomalous data. For this experiment we have modified the labels considering only two categories: anomalous or normal. We have reduced by so the 23 different types of attacks to one single class to separate from the normal data. This dataset has 18 million instances all of them labeled as normal or anomalous.

The models have been adapted to this problem by changing the specialisation module to build them as forecasting neural networks. The purpose of the training phase is to feed the architecture only with normal data. This procedure will force the model to learn only the normal data representation. Our final goal is to have an anomaly score for each of the instances. To achieve this goal the network receives a batch of data, from which it will predict the next 4 samples. As we have the real value of the next time steps we can calculate the error committed by the model, being this error the anomaly score of the predicted instances. As the network is trained with normal data the error while predicting normal instances will be low whereas the error predicting anomalous data will be higher as the neural network has only learned the normal data features. After all of the test data has been scored we label the top $K$ samples ordered by the anomaly score, labeling the samples with the highest anomaly scores as anomalous and the rest as normal where $K$ is the number of real labeled anomalies in the test data.

To train this model we have used the first $17.5$ million instances removing the anomalous samples, leaving at the end around $800K$ instances for training. To test the model the remaining $500K$ instances have been used. As batch size for feeding the model $4096$ instances are used to predict the next. For the training phase $150$ epochs are set, configuring as well an early stopping with patience $3$ and delta $0$.

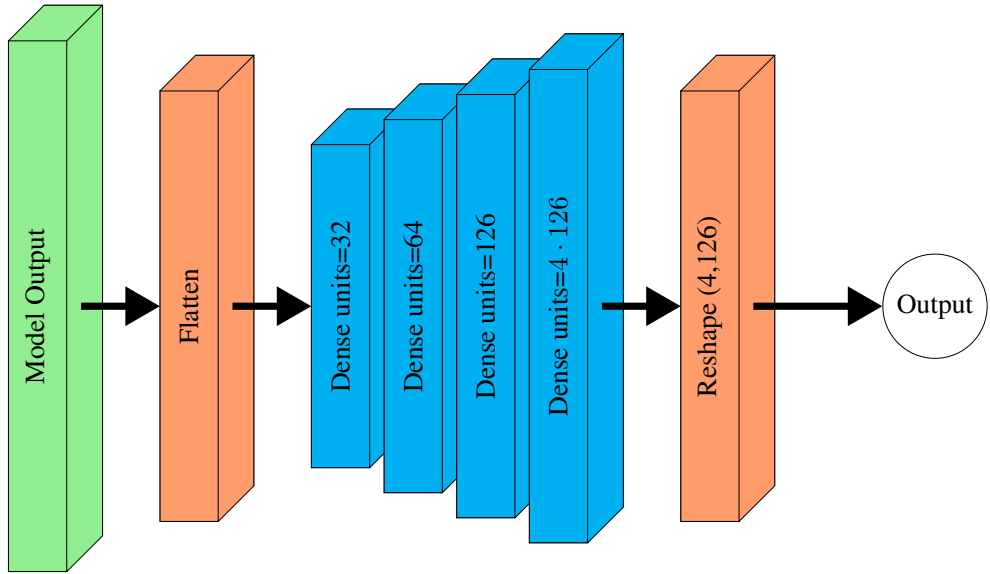
\begin{figure}[!hbt]
    \centering
    \resizebox{.8\linewidth}{!}{
        \begin{tikzpicture}
          \node[input](input){\rotatebox{90}{Model Output}};
          
          \node[conv,minimum height=6cm,right=1.3cm of input](flatten){\rotatebox{90}{Flatten}};
          
          \node[dense,minimum height=4.5cm,right=1.3cm of flatten](dense1){\rotatebox{90}{Dense units=32}};
          \node[dense,minimum height=5.2cm,right=0.2cm of dense1](dense2){\rotatebox{90}{Dense units=64}};
          \node[dense,minimum height=5.9cm,right=0.2cm of dense2](dense3){\rotatebox{90}{Dense units=126}};
          \node[dense,minimum height=6.6cm,right=0.2cm of dense3](dense4){\rotatebox{90}{Dense units=$4\cdot 126$}};
          
          \node[conv,minimum height=6cm,right=1.3cm of dense4](reshape){\rotatebox{90}{Reshape (4,126)}};
          
          \node[circle,draw,radius=1cm,right=2cm of reshape](out){Output};
        
          \draw [-triangle 60,link] ([xshift=0.2cm]input.east) -- (flatten.west);
          \draw [-triangle 60,link] ([xshift=0.2cm]flatten.east) -- (dense1.west);
          \draw [-triangle 60,link] ([xshift=0.2cm]dense4.east) -- (reshape.west);
          \draw [-triangle 60,link] ([xshift=0.2cm]reshape.east) -- (out.west);
        
        \end{tikzpicture}
     }
    \caption{Specialisation module for the anomaly detection task.}
    \label{fig:anomalies-specialisation-module}
\end{figure}

As specialisation module we have used the one described in \Cref{fig:anomalies-specialisation-module}. The first operation performed is a flattening of the ouput of the models, then four fully connected layers are placed to increase the dimensionality of the final output achieving $4\cdot 126$ features. Our dataset has $126$ features for each sample, therefore being this number $4$ complete instances of the dataset. Finally the output is reshaped in a matrix fashion to easily compare it to the real values.

\clearpage
\section*{Required Metadata}
\label{sec:required-metadata}

\section*{Current code version}
\label{sec:current-code-version}

\begin{table}[!hbt]
    \begin{tabular}{|l|p{6.5cm}|p{6.5cm}|}
        \hline
        \texttt{Nr.} & \texttt{Code metadata description} & \texttt{Please fill in this column} \\
        \hline
        C1 & Current code version & v1.0.3 \\
        \hline
        C2 & Permanent link to code/repository used of this code version & \url{https://github.com/ari-dasci/S-TSFE-DL} \\
        \hline
        C3 & Legal Code License   & GNU Affero General Public License v3.0 \\
        \hline
        C4 & Code versioning system used & Git \\
        \hline
        C5 & Software code languages, tools, and services used & Python 3, Keras, Tensorflow, PyTorch and PyTorch-Lightning \\
        \hline
        C6 & Compilation requirements, operating environments \& dependencies & OS-X, Unix-like or Microsoft Windows, a Python interpreter (3.7) and the following Python packages:
        pytorch-lightning, scikit-learn, tensorflow-gpu/tensorflow, torchmetrics, wfdb, obspy\\
        \hline
        C7 & If available Link to developer documentation/manual & \url{https://s-tsfe-dl.readthedocs.io/en/latest/} \\
        \hline
        C8 & Support email for questions & nacheteam@ugr.es\\
        \hline
    \end{tabular}
    \caption{Code metadata}
    \label{} 
\end{table}

\end{document}

%% file: tikz_header.tex
\usepackage{tikz}
\usetikzlibrary{backgrounds,calc,shadings,shapes.arrows,arrows,shapes.symbols,shadows,positioning,decorations.markings,decorations.pathreplacing,backgrounds,arrows.meta}

\usepackage{xcolor}
\definecolor{darkblue}{HTML}{1f4e79}
\definecolor{lightblue}{HTML}{00b0f0}
\definecolor{salmon}{HTML}{ff9c6b}
\definecolor{lightgreen}{HTML}{90ee90}

\usetikzlibrary{backgrounds,calc,shadings,shapes.arrows,arrows,shapes.symbols,shadows,positioning,decorations.markings,decorations.pathreplacing,backgrounds,arrows.meta}

\makeatletter
\pgfkeys{/pgf/.cd,
  parallelepiped offset x/.initial=2mm,
  parallelepiped offset y/.initial=2mm
}
\pgfdeclareshape{parallelepiped}
{
  \inheritsavedanchors[from=rectangle] 
  \inheritanchorborder[from=rectangle]
  \inheritanchor[from=rectangle]{north}
  \inheritanchor[from=rectangle]{north west}
  \inheritanchor[from=rectangle]{north east}
  \inheritanchor[from=rectangle]{center}
  \inheritanchor[from=rectangle]{west}
  \inheritanchor[from=rectangle]{east}
  \inheritanchor[from=rectangle]{mid}
  \inheritanchor[from=rectangle]{mid west}
  \inheritanchor[from=rectangle]{mid east}
  \inheritanchor[from=rectangle]{base}
  \inheritanchor[from=rectangle]{base west}
  \inheritanchor[from=rectangle]{base east}
  \inheritanchor[from=rectangle]{south}
  \inheritanchor[from=rectangle]{south west}
  \inheritanchor[from=rectangle]{south east}
  \backgroundpath{
    \southwest \pgf@xa=\pgf@x \pgf@ya=\pgf@y
    \northeast \pgf@xb=\pgf@x \pgf@yb=\pgf@y
    \pgfmathsetlength\pgfutil@tempdima{\pgfkeysvalueof{/pgf/parallelepiped
      offset x}}
    \pgfmathsetlength\pgfutil@tempdimb{\pgfkeysvalueof{/pgf/parallelepiped
      offset y}}
    \def\ppd@offset{\pgfpoint{\pgfutil@tempdima}{\pgfutil@tempdimb}}
    \pgfpathmoveto{\pgfqpoint{\pgf@xa}{\pgf@ya}}
    \pgfpathlineto{\pgfqpoint{\pgf@xb}{\pgf@ya}}
    \pgfpathlineto{\pgfqpoint{\pgf@xb}{\pgf@yb}}
    \pgfpathlineto{\pgfqpoint{\pgf@xa}{\pgf@yb}}
    \pgfpathclose
    \pgfpathmoveto{\pgfqpoint{\pgf@xb}{\pgf@ya}}
    \pgfpathlineto{\pgfpointadd{\pgfpoint{\pgf@xb}{\pgf@ya}}{\ppd@offset}}
    \pgfpathlineto{\pgfpointadd{\pgfpoint{\pgf@xb}{\pgf@yb}}{\ppd@offset}}
    \pgfpathlineto{\pgfpointadd{\pgfpoint{\pgf@xa}{\pgf@yb}}{\ppd@offset}}
    \pgfpathlineto{\pgfqpoint{\pgf@xa}{\pgf@yb}}
    \pgfpathmoveto{\pgfqpoint{\pgf@xb}{\pgf@yb}}
    \pgfpathlineto{\pgfpointadd{\pgfpoint{\pgf@xb}{\pgf@yb}}{\ppd@offset}}
  }
}
\makeatother

\tikzset{
  block/.style={
    parallelepiped,fill=white, draw,
    minimum width=2.4cm,
    minimum height=0.8cm,
    parallelepiped offset x=0.5cm,
    parallelepiped offset y=0.5cm,
    path picture={
      \draw[top color=darkblue,bottom color=darkblue]
        (path picture bounding box.south west) rectangle 
        (path picture bounding box.north east);
    },
    text=white,
  },
  conv/.style={
    parallelepiped,fill=white, draw,
    minimum width=0.8cm,
    minimum height=2.4cm,
    parallelepiped offset x=0.5cm,
    parallelepiped offset y=0.5cm,
    path picture={
      \draw[top color=salmon,bottom color=salmon]
        (path picture bounding box.south west) rectangle 
        (path picture bounding box.north east);
    },
    text=black,
  },
  input/.style={
    parallelepiped,fill=white, draw,
    minimum width=0.8cm,
    minimum height=7.4cm,
    parallelepiped offset x=0.5cm,
    parallelepiped offset y=0.5cm,
    path picture={
      \draw[top color=lightgreen,bottom color=lightgreen]
        (path picture bounding box.south west) rectangle 
        (path picture bounding box.north east);
    },
    text=black,
  },
  dense/.style={
    parallelepiped,fill=white, draw,
    minimum width=0.8cm,
    minimum height=2.4cm,
    parallelepiped offset x=0.5cm,
    parallelepiped offset y=0.5cm,
    path picture={
      \draw[top color=lightblue,bottom color=lightblue]
        (path picture bounding box.south west) rectangle 
        (path picture bounding box.north east);
    },
    text=black,
  },
  link/.style={
    color=black,
    line width=1mm,
  },
  cbrace/.style={
    decorate,
    decoration={brace,amplitude=0.5cm,mirror},
    line width=1mm,
  },
}